\documentclass[10pt,twocolumn,letterpaper]{article}

\usepackage[pagenumbers]{cvpr} 

\usepackage[T1]{fontenc}
\usepackage{multirow}
\usepackage{array}
\usepackage{colortbl}
\usepackage{pifont}
\definecolor{scoregraylight}{gray}{0.88}
\definecolor{scoregraydark}{gray}{0.88}
\definecolor{groupfogblue}{RGB}{226,235,240}
\newcommand{\cmark}{\ding{51}}
\newcommand{\xmark}{\ding{55}}

\def\combinedbuild{}

\definecolor{cvprblue}{rgb}{0.21,0.49,0.74}
\usepackage[breaklinks,colorlinks,allcolors=cvprblue]{hyperref}

\def\confName{CVPR}

\title{DriveWLA: Unifying Future Dynamics and Scene Semantics through World-Language-Action Modeling for Autonomous Driving}

\author{
  \textbf{Shuzheng Wang$^{1,2}$, Haoxuan Li$^{1,2}$, Shengzhen Li$^{1,2}$, Chenxi Wu$^{1,2}$}\\
  \textbf{Yajing Sun$^{2}$, Lu Hou$^{2}$, Wenyu Liu$^{1}$, Xinggang Wang$^{1,\text{\normalfont\ding{41}}}$}\\[3pt]
  \small $^{1}$Huazhong University of Science and Technology\quad
  $^{2}$Yinwang Intelligent Technology Co. Ltd.\\
  \small $^{\text{\normalfont\ding{41}}}$Corresponding author.\quad
  \texttt{\{shuzwang,xgwang\}@hust.edu.cn}
}

\newdimen\authorblockshift
\makeatletter
\def\@maketitle{
   \newpage
   \null
   \setbox\@tempboxa\vbox{\large\lineskip .5em
     \begin{tabular}[t]{c}\@author\end{tabular}}%
   \authorblockshift=\dimexpr\ht\@tempboxa+\dp\@tempboxa\relax
   \setbox\@tempboxa\vbox{\large\lineskip .5em
     \begin{tabular}[t]{c}Anonymous \confName~submission\end{tabular}}%
   \advance\authorblockshift by -\dimexpr\ht\@tempboxa+\dp\@tempboxa\relax
   \advance\authorblockshift by 6pt
   \typeout{DriveWLA title block shifted up by \the\authorblockshift}%
   \vskip .375in
   \vskip -\authorblockshift
   \vskip -58pt
   \begin{center}
      {\Large \bf \@title \par}
      \vspace*{24pt}{
        \large
        \lineskip .5em
        \begin{tabular}[t]{c}
          \@author
        \end{tabular}
        \par
      }
      \vskip .5em
      \vspace*{12pt}
   \end{center}
}
\makeatother

\hypersetup{
  pdftitle={DriveWLA: Unifying Future Dynamics and Scene Semantics through World-Language-Action Modeling for Autonomous Driving},
  pdfauthor={Shuzheng Wang, Haoxuan Li, Shengzhen Li, Chenxi Wu, Yajing Sun, Lu Hou, Wenyu Liu, Xinggang Wang},
}

\begin{document}
\maketitle
\begin{abstract}
Vision-Language Models (VLMs) and world models have demonstrated strong capabilities in semantic understanding and future scene evolution for autonomous driving. However, effectively integrating both capabilities for trajectory generation remains challenging. In this work, we propose \textbf{DriveWLA}, a novel unified World-Language-Action model based on Mixture-of-Transformers (MoT) that jointly models semantic understanding, future scene dynamics, and trajectory generation. DriveWLA consists of three experts: a World Expert for future prediction, a Language Expert for semantic understanding, and an Action Expert for trajectory generation. 
We further introduce Cross-Expert Attention, which enables controlled information flow across experts while aligning semantic and dynamic representations. 
To improve trajectory generation beyond imitation learning, we apply Flow-GRPO reinforcement learning post-training to refine the Action Expert. 
Extensive experiments show that DriveWLA achieves superior performance in both open-loop and closed-loop evaluation. 
DriveWLA achieves 92.3 PDMS on NAVSIM v1 and 90.1 EPDMS on NAVSIM v2, demonstrating strong performance among VLM-based and world-model-based planners. 
Without specific fine-tuning, DriveWLA further demonstrates strong zero-shot generalization to HUGSIM, nuScenes, and Bench2Drive. 
These results validate the effectiveness of our unified World-Language-Action modeling, establishing DriveWLA as a powerful and generalizable framework for autonomous driving.

\end{abstract}

\begin{figure}[t]
\centering
\includegraphics[width=\linewidth]{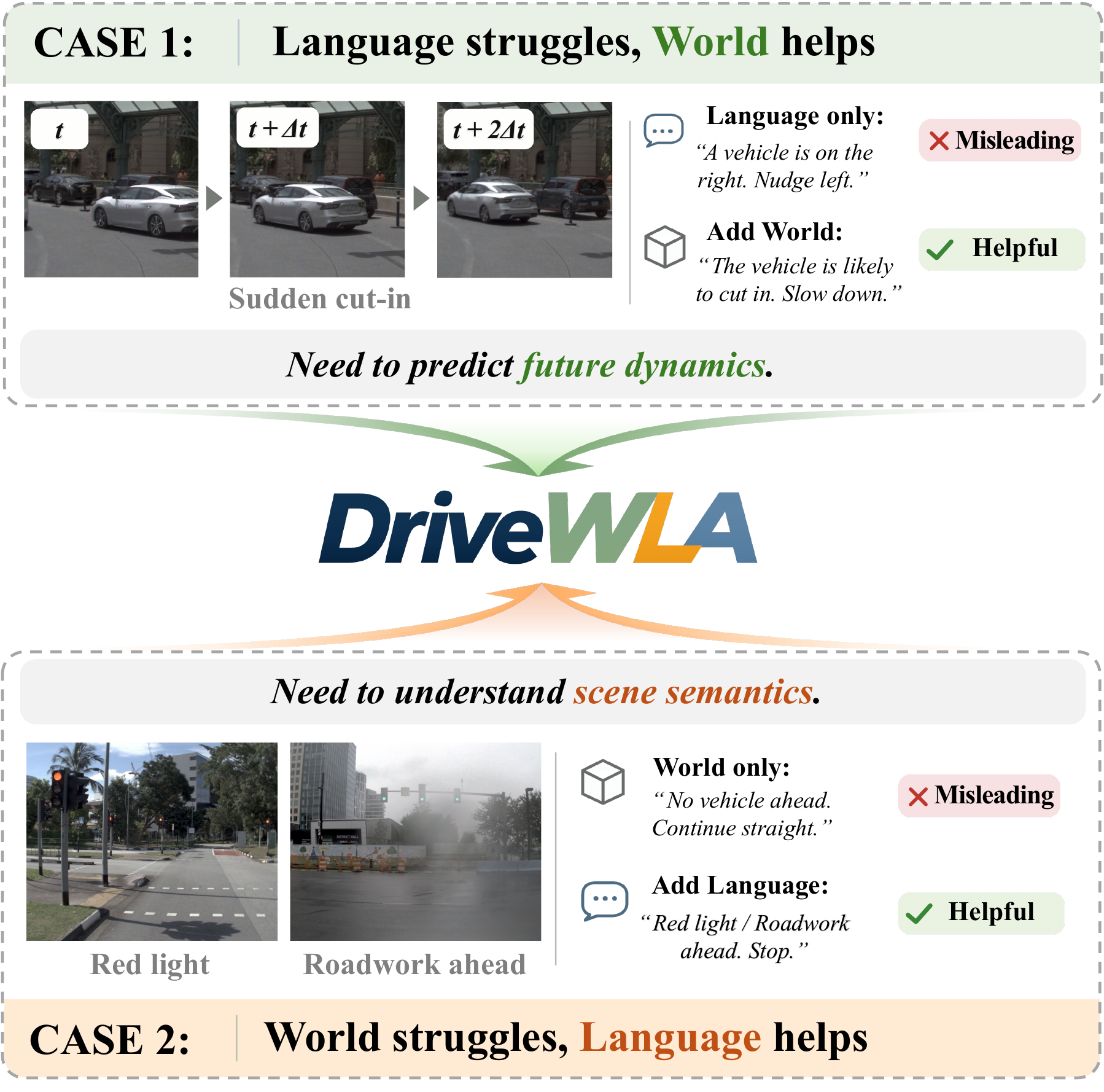}
\caption{\textbf{Complementary roles of language and world models.} Future dynamics help anticipate vehicle interactions, while scene semantics provide traffic-rule context. DriveWLA integrates both capabilities for trajectory planning.}
\label{fig:intro}
\end{figure}

\section{Introduction}

Reliable autonomous driving requires three core capabilities: scene understanding, future prediction, and trajectory planning~\citep{hu2023gaia,hu2023planning,wang2024drivedreamer}. Human drivers naturally combine current scene understanding with future prediction to make appropriate driving decisions~\citep{engstrom2018great,wolfe2019predicting}. This motivates a key challenge: how to jointly leverage semantic understanding and future dynamics for trajectory planning.

Vision-Language Models (VLMs) have recently demonstrated strong capabilities in semantic understanding and reasoning~\citep{shao2024lmdrive,zhu2025internvl3}. 
By leveraging large-scale vision-language pretraining and driving-specific adaptation, VLM-based driving systems can understand complex traffic scenarios and predict driving behaviors~\citep{li2026sgdrive,xiong2026recogdrive,zhou2026autovla}. 
However, VLMs inherently lack explicit mechanisms for future scene evolution, limiting their ability to predict future dynamics required for trajectory planning.

In contrast, world models can learn future scene evolution from large-scale driving videos and provide future-aware priors beyond current observations~\citep{hu2023gaia,zhao2025drivedreamer,wang2024driving}. 
However, future prediction alone is insufficient for planning, as the model must also understand the current scene to determine which future changes are relevant to driving. Therefore, semantic understanding and future modeling are complementary for trajectory planning (Fig.~\ref{fig:intro}).


Recent studies have explored integrating VLMs with world models~\citep{li2026drivevla,shi2026drivewam,zhan2026brainwam}. However, most methods introduce one capability as an enhancement to another framework. 
For example, DriveVLA-W0~\citep{li2026drivevla} adds future-image prediction as an auxiliary objective for VLA training, while DriveWAM~\citep{shi2026drivewam} uses VLM features to condition video and action generation. Although these designs bring semantic understanding and future prediction into the same system, they do not explicitly model their interaction for trajectory planning. As a result, the system cannot unlock the full potential of their complementary information for driving decisions due to insufficient refinement and alignment between the semantic and dynamic representations.

To address this challenge, we propose DriveWLA, the first unified World-Language-Action framework for autonomous driving that successfully integrates three specialized experts through controlled information flow: a World Expert for future prediction, a Language Expert for semantic understanding, and an Action Expert for trajectory generation. 
We employ Cross-Expert Attention to explicitly control information exchange among the three experts. With this interaction mechanism, all experts are jointly trained under trajectory supervision during imitation learning, aligning the pretrained language and world representations with trajectory generation. We then refine the driving policy with an improved Flow-GRPO~\citep{liu2026flow}, which uses factorized advantages and variance-based shrinkage for stable reward optimization. This reinforcement
learning stage enables the model to explore better trajectories beyond expert demonstrations.

DriveWLA achieves 92.3 PDMS on NAVSIM v1~\citep{dauner2024navsim}, establishing state-of-the-art (SOTA) performance among recent VLM-based and world-model-based planners.
We further evaluate its zero-shot capability on representative open-loop and closed-loop benchmarks without fine-tuning. DriveWLA achieves 41.47 HD-Score on HUGSIM~\citep{zhou2025hugsim} and maintains low collision rates on nuScenes~\citep{caesar2020nuscenes} and Bench2Drive~\citep{jia2024bench2drive}, demonstrating strong generalization of the proposed unified architecture. Our contributions can be summarized as follows:
\begin{itemize}
\setlength{\itemsep}{1pt}
\setlength{\topsep}{2pt}

\item We introduce DriveWLA, a novel unified World-Language-Action framework that jointly models semantic understanding, future dynamics prediction, and trajectory generation through three specialized experts.

\item We propose Cross-Expert Attention, a controllable mechanism that explicitly directs information flow across experts, enabling both efficient interaction and representation alignment.

\item DriveWLA establishes a new state-of-the-art on NAVSIM v1 among VLM-based and world-model-based planners. Furthermore, it generalizes zero-shot to diverse open-loop and closed-loop benchmarks, achieving competitive driving performance and safety.

\end{itemize}

\section{Related Work}

\begin{figure*}[t]
\centering
\includegraphics[width=\linewidth]{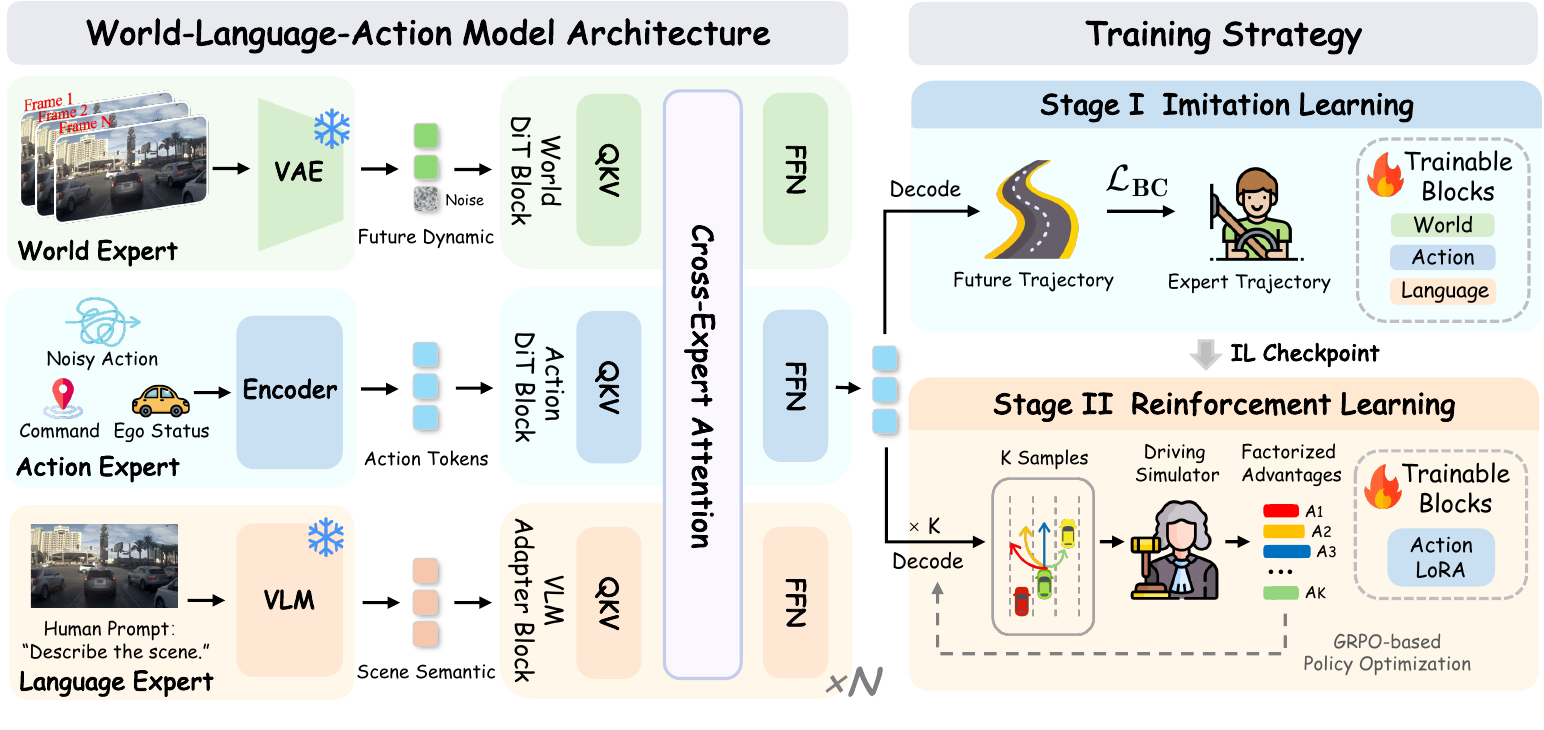}
\caption{\textbf{Model architecture and training strategy.} DriveWLA unifies World, Language, and Action Experts through Cross-Expert Attention. The training strategy consists of trajectory imitation learning, which jointly optimizes all experts, followed by reinforcement learning refinement of the Action Expert.}
\label{fig:overview}
\end{figure*}

\noindent\textbf{Vision-Language Models and world models for autonomous driving.}
VLM-based driving systems mainly leverage VLMs' strong scene understanding and reasoning capabilities through explicit textual reasoning or latent representations~\citep{hwang2024emma,tian2024drivevlm,zhou2026autovla,zhou2026opendrivevla}. 
ReCogDrive~\citep{xiong2026recogdrive} incorporates cognitive representations into a diffusion planner, while SGDrive~\citep{li2026sgdrive} learns structured spatiotemporal representations for trajectory prediction. 
World models, in contrast, learn predictive representations of future scene evolution from driving videos and provide dynamic priors for planning~\citep{hu2023gaia,wang2024driving,xia2026drivelaw,xia2026reworld,zhao2026simwam}. 
World-model-based approaches generally follow cascaded or joint formulations. 
Cascaded methods, such as DriveLaW~\citep{xia2026drivelaw}, predict future representations before using them for trajectory planning. 
Joint methods instead model future dynamics and action generation within a unified process: SimWAM~\citep{zhao2026simwam} jointly trains video and action experts without explicitly feeding predicted future representations into action generation, while DriveVA~\citep{liu2026driveva} models future video latents and actions within a shared diffusion framework.

\noindent\textbf{Methods to integrate VLMs and world models.}
Recent works explore integrating semantic understanding from VLMs with future dynamics from world models~\citep{li2026drivevla,shi2026drivewam,liu2026driveworld}. 
These methods mainly follow two strategies: introducing complementary capabilities through auxiliary objectives or using representations from one paradigm to condition another. 
DriveVLA-W0~\citep{li2026drivevla} introduces future-image prediction as an auxiliary objective for VLA training, while DriveWAM~\citep{shi2026drivewam} and DriveWorld-VLA~\citep{liu2026driveworld} use semantic and dynamic representations, respectively, as conditioning information to augment their existing architectures. 
In contrast, DriveWLA jointly models World, Language, and Action Experts through Cross-Expert Attention, enabling continuous information exchange and representation alignment.

\noindent\textbf{Reinforcement learning for autonomous driving.}
Behavior cloning learns driving policies from expert trajectories, whereas reinforcement learning further improves them using task-level driving rewards~\citep{gao2026rad,zhang2025carplanner,yang2026raw2drive,huang2025gen}. 
RAD~\citep{gao2026rad} applies reinforcement learning to end-to-end driving models in realistic simulations. 
Recent works extend reinforcement learning to generative trajectory planners~\citep{huang2025gen,xiong2026recogdrive,xu2026eponav2}. 
ReCogDrive~\citep{xiong2026recogdrive} introduces DiffGRPO for diffusion-based trajectory planning, while EponaV2~\citep{xu2026eponav2} adapts GRPO to flow-matching-based planners~\citep{liu2026flow}. 
DriveWLA further incorporates factorized advantages and variance-based shrinkage to preserve multi-dimensional driving reward signals during policy optimization.

\section{Method}
\begingroup
\raggedbottom
\setlength{\abovedisplayskip}{4pt}
\setlength{\belowdisplayskip}{4pt}
\setlength{\abovedisplayshortskip}{2pt}
\setlength{\belowdisplayshortskip}{2pt}
\setlength{\textfloatsep}{12pt plus 2pt minus 2pt}
\setlength{\dbltextfloatsep}{12pt plus 2pt minus 2pt}
\widowpenalty=10000
\displaywidowpenalty=0
\predisplaypenalty=0
\postdisplaypenalty=0
\clubpenalty=150
\makeatletter\@clubpenalty=150\makeatother
\label{sec:method}
\noindent\textbf{Problem Definition.}
\label{sec:problem}
Autonomous driving aims to generate a safe and smooth future trajectory from camera observations $\mathcal I_{\rm cam}$, historical trajectories $H_{\rm traj}$, navigation commands $\mathcal C_{\rm nav}$, and ego status $\mathcal S_{\rm ego}$~\citep{codevilla2018end,hu2023planning}. For an end-to-end driving system $\Phi$, this is formulated as
\begin{equation}
\mathrm{T_{traj}}=\Phi(\mathcal I_{\rm cam},H_{\rm traj},\mathcal C_{\rm nav},\mathcal S_{\rm ego}),
\label{eq:planning}
\end{equation}
where $\mathrm{T_{traj}}=\{(x_t,y_t,\theta_t)\}_{t=1}^{T}\in\mathbb R^{T\times3}$ denotes the predicted ego trajectory over $T$ future steps, with $(x_t,y_t)$ and $\theta_t$ representing the position and heading in the current ego-centric frame.

In this section, we present the overall architecture and training paradigm of DriveWLA, as shown in Fig.~\ref{fig:overview}.
First, we introduce the three specialized experts and detail their respective architectures and roles in trajectory planning.
Second, we detail Cross-Expert Attention, which enables controlled information exchange and representation alignment among the three experts at every transformer block. 
Finally, we introduce our improved version of Flow-GRPO~\citep{liu2026flow}, which uses factorized advantage estimation to refine the Action Expert.

\subsection{Model Architecture}
\label{sec:experts}
\noindent
\textbf{World Expert.}
The World Expert is initialized from a video generation model pretrained on large-scale driving videos and consists of a spatiotemporal VAE, a text encoder, and an $N$-block Diffusion Transformer (DiT)~\citep{xia2026drivelaw,hacohen2024ltx,peebles2023scalable}. Given the current and historical camera observations $\mathcal I_{\rm cam}$, the VAE encodes the observed frames into spatiotemporal \mbox{latent $Z_{\rm obs}$}, while the future-frame latent $Z_{\rm fut}$ is initialized from Gaussian noise for future prediction:
\begin{equation}
\begin{gathered}
Z_{\rm obs}=E_{\rm VAE}(\mathcal I_{\rm cam}),\quad
Z_{\rm fut}\sim\mathcal N(\mathbf 0,\mathbf I),\\
Z_W=[Z_{\rm obs};Z_{\rm fut}].
\end{gathered}
\label{eq:world-input}
\end{equation}
Here, $[Z_{\rm obs};Z_{\rm fut}]$ denotes concatenation of the observed and future latents along the temporal dimension. Conditioned on the encoded generation instruction, the World blocks produce layer-wise predictive representations of future scene evolution in a single forward pass, which are directly used for planning without iterative video sampling or future-frame decoding.

\noindent\textbf{Language Expert and VLM Adapter.}
The Language Expert consists of a frozen driving-pretrained VLM and an $N$-block VLM Adapter Transformer. We provide the VLM with camera observations $\mathcal I_{\rm cam}$ and a driving prompt $P$ constructed from ego-motion information, navigation commands, and human instructions. After processing the visual-textual inputs, we directly extract the final-layer hidden states of valid input tokens:
\begin{equation}
H_L=\mathcal{F}_{\rm VLM}^{\rm last}(\mathcal I_{\rm cam},P),
\label{eq:language}
\end{equation}
where $\mathcal{F}_{\rm VLM}^{\rm last}(\cdot)$ denotes the mapping from visual-textual inputs to the final-layer hidden representations of the VLM. We do not perform autoregressive decoding for textual outputs; instead, these hidden states are directly used as the semantic representations and further processed by the VLM Adapter to obtain planning-oriented features.

\noindent\textbf{Action Expert.}
The Action Expert is a lightweight $N$-block DiT~\citep{peebles2023scalable} that generates future trajectories with a flow-matching policy~\citep{lipman2022flow}. Given a noisy trajectory state $a_\tau$, flow time $\tau$, and the driving condition $\mathcal C_d$ composed of ego information and navigation commands, the Action Expert predicts the velocity field $v_{\theta}(a_\tau,\tau,\mathcal C_d)$.

\par

\noindent\textbf{Training Objective.}
During imitation learning, the World Expert, VLM Adapter, and Action Expert are jointly optimized with the flow-matching behavior cloning objective. Specifically, let $a_0$ denote the expert trajectory and $a_1\sim\mathcal N(\mathbf 0,\mathbf I)$ denote Gaussian noise. The interpolated trajectory state is defined as:
$a_\tau=(1-\tau)a_0+\tau a_1$.
The flow-matching behavior cloning objective is:
\begin{equation}
\mathcal L_{\rm BC}
=
\mathbb E
\left[
\left\|
v_{\theta}(a_\tau,\tau,\mathcal C_d)-(a_1-a_0)
\right\|_2^2
\right].
\label{eq:action-bc}
\end{equation}

\subsection{Cross-Expert Attention}
\label{sec:amja}

\begin{figure}[t]
\centering
\includegraphics[width=0.85\linewidth]{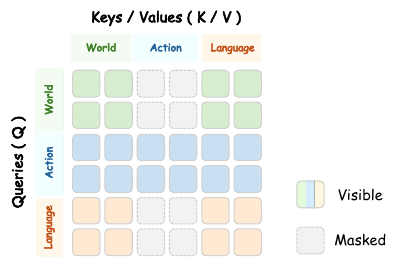}
\caption{Query-Key/Value Visibility in Cross-Expert Attention.}
\label{fig:attention-mask}
\end{figure}

Although the World and Language Experts provide complementary features, effectively integrating them remains challenging. Since their representations are shaped by different learning objectives, naive fusion may lead to representation conflicts. To address this issue, we introduce Cross-Expert Attention within each transformer block, which enables controlled information exchange and progressive alignment among the three experts.

Inspired by the mixture-of-transformers (MoT) architecture~\citep{liang2024mixture}, Cross-Expert Attention allows the three experts to exchange information while maintaining their own transformer blocks. The inputs of the three experts are first projected into a shared dimension $d$, forming the initial inputs $W^0$, $A^0$, and $L^0$. At block $\ell$, the hidden states of each expert $E^\ell$, where $E\in\{W,L,A\}$, are mapped to queries, keys, and values using their corresponding projections:
\begin{equation}
Q_E^\ell=E^\ell W_{Q,E}^\ell,\;
K_E^\ell=E^\ell W_{K,E}^\ell,\;
V_E^\ell=E^\ell W_{V,E}^\ell .
\label{eq:cea-qkv}
\end{equation}
The query, key, and value matrices from the three experts are then concatenated as $Q^\ell$, $K^\ell$, $V^\ell$, and processed by masked self-attention:
\begin{equation}
O^\ell=\operatorname{softmax}\!\left(
\frac{Q^\ell(K^\ell)^\top}{\sqrt{d}}+M
\right)V^\ell .
\label{eq:cea-attention}
\end{equation}
\noindent Here, $M$ defines the visibility among the three experts, as illustrated in Fig.~\ref{fig:attention-mask}. The World and Language Experts have bidirectional visibility, allowing their representations to mutually exchange information. The Action Expert attends to both representations for trajectory generation, while its representations are prevented from flowing back to the World and Language Experts.

The attention outputs are split back into the corresponding expert branches, and each output is further processed by its expert-specific output projection, residual connection, normalization, and feed-forward transformation.

\subsection{Policy Refinement with Flow-GRPO}
\begin{table*}[t]
\captionsetup{position=top,skip=4pt}
\caption{\textbf{Comparison of planners based on different paradigms on NAVSIM v1 \texttt{navtest}.} Bold values denote the best result for each metric. The upward arrow ($\uparrow$) indicates that higher values are better.}
\label{tab:navsim-v1}
\centering
\begingroup
\small
\setlength{\tabcolsep}{0pt}
\setlength{\arrayrulewidth}{0.3pt}
\renewcommand{\arraystretch}{1.10}
\begin{tabular}{@{}>{\raggedright\arraybackslash}p{\dimexpr0.19\linewidth-0.3pt\relax}|>{\centering\arraybackslash}p{\dimexpr0.18\linewidth-0.3pt\relax}>{\centering\arraybackslash}p{\dimexpr0.15\linewidth-0.3pt\relax}|>{\centering\arraybackslash}p{0.068333\linewidth}>{\centering\arraybackslash}p{0.068333\linewidth}>{\centering\arraybackslash}p{0.068333\linewidth}>{\centering\arraybackslash}p{0.068333\linewidth}>{\centering\arraybackslash}p{0.068333\linewidth}@{\hspace{0.023335\linewidth}}|>{\columncolor{scoregraylight}[-3pt][0pt]\centering\arraybackslash}p{0.115\linewidth}@{}}
\toprule
\textbf{Method} & \textbf{VLM Backbone} & \textbf{Video Backbone} & \textbf{NC}$\uparrow$ & \textbf{DAC}$\uparrow$ & \textbf{TTC}$\uparrow$ & \textbf{Comf.}$\uparrow$ & \textbf{EP}$\uparrow$ & \textbf{PDMS}$\uparrow$\\
\midrule
\rowcolor{groupfogblue!45}[0pt][0pt]\multicolumn{9}{@{}l@{}}{\emph{VLM-based planners}}\\
ReCogDrive~\citep{xiong2026recogdrive} & InternVL3-2B & -- & 97.9 & 97.3 & 94.9 & \textbf{100.0} & \textbf{87.3} & 90.8\\
SGDrive~\citep{li2026sgdrive} & InternVL3-2B & -- & 98.6 & 97.8 & 96.2 & \textbf{100.0} & 85.8 & 91.1\\
Qwen-Drive-1.0~\citep{zhou2026qwen} & Qwen3.5-4B & -- & 98.6 & 98.2 & 95.9 & \textbf{100.0} & 84.8 & 90.7\\
\midrule
\rowcolor{groupfogblue!45}[0pt][0pt]\multicolumn{9}{@{}l@{}}{\emph{World-Model-based planners}}\\
DriveLaW~\citep{xia2026drivelaw} & -- & LTX-Video-2B & 99.0 & 97.1 & 96.7 & \textbf{100.0} & 81.3 & 89.1\\
ReWorld~\citep{xia2026reworld} & -- & LTX-Video-2B & \textbf{99.1} & 98.2 & \textbf{97.7} & 99.8 & 82.0 & 90.4\\
SimWAM~\citep{zhao2026simwam} & -- & Wan2.2-5B & 98.4 & 98.7 & 95.5 & \textbf{100.0} & 86.4 & 91.5\\
\midrule
\rowcolor{groupfogblue!45}[0pt][0pt]\multicolumn{9}{@{}l@{}}{\emph{VLM--WM--based planners}}\\
DriveVLA-W0~\citep{li2026drivevla} & \multicolumn{2}{c|}{Emu3-8B (unified)} & 98.7 & \textbf{99.1} & 95.3 & 99.3 & 83.3 & 90.2\\
DriveWAM~\citep{shi2026drivewam} & Qwen3-VL-8B & Wan2.2-5B & 98.3 & 98.1 & 95.2 & \textbf{100.0} & 84.3 & 90.1\\
BrainWAM~\citep{zhan2026brainwam} & Qwen3-VL-4B & Wan2.2-5B & 98.1 & 97.5 & 94.9 & \textbf{100.0} & 83.8 & 89.5\\
DriveWorld-VLA~\citep{liu2026driveworld} & InternVL3-2B & ResNet-34 & \textbf{99.1} & 98.2 & 96.1 & \textbf{100.0} & 85.9 & 91.3\\
\midrule
\textbf{DriveWLA (Ours)} & InternVL3-2B & LTX-Video-2B & \textbf{99.1} & 98.8 & 97.0 & \textbf{100.0} & 86.4 & \textbf{92.3}\\
\bottomrule
\end{tabular}
\endgroup
\end{table*}

\begin{table*}[t]
\captionsetup{position=top,skip=4pt}
\caption{\textbf{Comparison on the NAVSIM v2 \texttt{navtest}} split.
All results use the corrected official evaluator with human-reference filtering.}
\label{tab:navsim-v2}
\centering
\begingroup
\small
\setlength{\tabcolsep}{0pt}
\setlength{\arrayrulewidth}{0.3pt}
\renewcommand{\arraystretch}{1.10}
\begin{tabular}{@{}>{\raggedright\arraybackslash}p{\dimexpr0.19\linewidth-0.6pt\relax}|@{\hspace{0.01\linewidth}}>{\centering\arraybackslash}p{0.075\linewidth}>{\centering\arraybackslash}p{0.075\linewidth}>{\centering\arraybackslash}p{0.075\linewidth}>{\centering\arraybackslash}p{0.075\linewidth}>{\centering\arraybackslash}p{0.075\linewidth}>{\centering\arraybackslash}p{0.075\linewidth}>{\centering\arraybackslash}p{0.075\linewidth}>{\centering\arraybackslash}p{0.075\linewidth}>{\centering\arraybackslash}p{0.075\linewidth}@{\hspace{0.01\linewidth}}|>{\columncolor{scoregraydark}[-3pt][0pt]\centering\arraybackslash}p{0.115\linewidth}@{}}
\toprule
\textbf{Method} & \textbf{NC}$\uparrow$ & \textbf{DAC}$\uparrow$ & \textbf{DDC}$\uparrow$ & \textbf{TLC}$\uparrow$ & \textbf{EP}$\uparrow$ & \textbf{TTC}$\uparrow$ & \textbf{LK}$\uparrow$ & \textbf{HC}$\uparrow$ & \textbf{EC}$\uparrow$ & \textbf{EPDMS}$\uparrow$\\
\midrule
DiffusionDrive~\citep{liao2025diffusiondrive} & 98.2 & 95.9 & 99.4 & 99.8 & 87.5 & 97.3 & 96.8 & 98.3 & 87.7 & 84.5\\
DiffusionDriveV2~\citep{zou2025diffusiondrivev2} & 97.7 & 96.6 & 99.2 & 99.8 & \textbf{88.9} & 97.2 & 96.0 & 97.8 & \textbf{91.0} & 87.5\\
DriveWorld-VLA~\citep{liu2026driveworld} & 98.6 & \textbf{99.1} & 99.6 & 99.8 & 87.4 & 97.9 & 97.0 & 97.8 & 78.6 & 86.8\\
Latent-WAM~\citep{wang2026latent} & 98.1 & 97.3 & 99.6 & 99.8 & 87.7 & 97.3 & 97.6 & 98.1 & 87.3 & 89.3\\
CoWorld-VLA~\citep{huang2026coworld} & \textbf{99.1} & 97.0 & 99.6 & \textbf{99.9} & 87.9 & 98.5 & 97.7 & 98.2 & 86.2 & 90.0\\
DriveFuture~\citep{hong2026drivefuture} & 98.8 & \textbf{99.1} & 99.6 & \textbf{99.9} & 86.6 & 98.4 & 96.4 & 98.3 & 74.8 & 89.9\\
\midrule
\textbf{DriveWLA (Ours)} & 98.8 & 97.8 & \textbf{99.7} & \textbf{99.9} & 87.4 & \textbf{98.8} & \textbf{97.8} & \textbf{98.4} & 82.5 & \textbf{90.1}\\
\bottomrule
\end{tabular}
\endgroup
\end{table*}

\begin{table*}[t]
\captionsetup{position=top,skip=4pt}
\centering
\caption{\textbf{Zero-shot closed-loop evaluation on HUGSIM.} We report driving metrics over all 436 scenarios and HD-Score across different difficulty levels, together with the overall average.}
\label{tab:hugsim}
\small
\setlength{\tabcolsep}{3pt}
\renewcommand{\arraystretch}{1.10}
\begin{tabular*}{\linewidth}{@{\extracolsep{\fill}}lcccccccccc>{\columncolor{scoregraylight}[\tabcolsep][0pt]}c@{}}
\toprule
\multirow{2}{*}{\textbf{Method}} & \multicolumn{6}{c}{\textbf{All 436 Scenarios}} & \multicolumn{5}{c}{\textbf{HD-Score}} \\
\cmidrule(lr){2-7}\cmidrule(l){8-12}
& NC$\uparrow$ & DAC$\uparrow$ & TTC$\uparrow$ & Comf.$\uparrow$ & PDMS$\uparrow$ & RC$\uparrow$ & Easy$\uparrow$ & Medium$\uparrow$ & Hard$\uparrow$ & Extreme$\uparrow$ & Avg.$\uparrow$ \\
\midrule
VAD~\citep{jiang2023vad} & 41.20 & 90.30 & 28.00 & \textbf{95.30} & 28.30 & 30.10 & 42.00 & 8.50 & 7.70 & 6.30 & 13.90 \\
UniAD~\citep{hu2023planning} & 65.60 & 93.20 & 51.60 & 66.30 & 49.40 & 43.80 & 64.00 & 37.20 & 21.00 & 6.30 & 31.20 \\
LTF~\citep{chitta2022transfuser,zhou2025hugsim} & 44.30 & 92.80 & 37.50 & 94.80 & 36.50 & 38.00 & 66.10 & 15.50 & 12.00 & 11.70 & 23.10 \\
DrivoR~\citep{kirby2026driving} & 52.20 & \textbf{95.60} & 46.20 & 93.90 & 44.80 & 47.20 & \textbf{78.00} & 29.10 & 20.00 & 14.10 & 32.50 \\
\midrule
\textbf{DriveWLA (Ours)} & \textbf{65.80} & 95.06 & \textbf{59.27} & 75.78 & \textbf{54.92} & \textbf{54.99} & 73.87 & \textbf{47.91} & \textbf{31.41} & \textbf{15.87} & \textbf{41.47} \\
\bottomrule
\end{tabular*}
\end{table*}

\begin{table*}[t]
\captionsetup{position=top,skip=4pt}
\centering
\caption{\textbf{Zero-shot open-loop evaluation on nuScenes and Bench2Drive.} We report L2 trajectory error (m) and collision rate (\%) over 1--3 seconds and their averages. ``Fine-tune'' indicates target-dataset adaptation. Bold denotes the best result for each metric within each dataset.}
\label{tab:transfer}
\small
\setlength{\tabcolsep}{3pt}
\renewcommand{\arraystretch}{1.10}
\begin{tabular*}{\linewidth}{@{\extracolsep{\fill}}ll c c c c >{\columncolor{scoregraylight}}c c c c >{\columncolor{scoregraylight}[\tabcolsep][0pt]}c@{}}
\toprule
\multirow{2}{*}{\textbf{Dataset}} & \multirow{2}{*}{\textbf{Method}} & \multirow{2}{*}{\textbf{Fine-tune}} & \multicolumn{4}{c}{\textbf{L2 (m)}$\downarrow$} & \multicolumn{4}{c}{\textbf{Collision (\%)} $\downarrow$}\\
\cmidrule(lr){4-7}\cmidrule(lr){8-11}
& & & 1s & 2s & 3s & Avg. & 1s & 2s & 3s & Avg.\\
\midrule
\multirow{6}{*}{nuScenes}
& UniAD~\citep{hu2023planning} & \textcolor{green!60!black}{\cmark} & 0.48 & 0.96 & 1.65 & 1.03 & 0.05 & 0.17 & 0.71 & 0.31\\
& OpenDriveVLA~\citep{zhou2026opendrivevla} & \textcolor{green!60!black}{\cmark} & \textbf{0.19} & \textbf{0.58} & \textbf{1.24} & \textbf{0.67} & \textbf{0.02} & 0.18 & 0.70 & 0.30\\
& UniDriveVLA-L~\citep{li2026unidrivevla} & \textcolor{green!60!black}{\cmark} & 0.30 & 0.66 & 1.25 & 0.74 & \textbf{0.02} & 0.18 & 0.40 & 0.20\\
& DriveVLA-W0~\citep{li2026drivevla} & \textcolor{red!75!black}{\xmark} & 0.43 & 1.26 & 2.60 & 1.43 & 0.22 & 0.66 & 1.42 & 0.77\\
& PWM~\citep{zhao2026forecasting} & \textcolor{red!75!black}{\xmark} & 2.06 & 3.91 & 6.00 & 3.99 & 0.12 & 0.15 & 0.86 & 0.36\\
\cmidrule(lr){2-11}
& \textbf{DriveWLA (Ours)} & \textcolor{red!75!black}{\xmark} & 0.38 & 0.83 & 1.60 & 0.94 & 0.03 & \textbf{0.03} & \textbf{0.20} & \textbf{0.09}\\
\midrule
\multirow{5}{*}{Bench2Drive}
& DriveVLA-W0~\citep{li2026drivevla} & \textcolor{red!75!black}{\xmark} & 1.01 & 2.77 & 5.22 & 3.00 & 1.49 & 2.53 & 3.53 & 2.52\\
& PWM~\citep{zhao2026forecasting} & \textcolor{red!75!black}{\xmark} & 1.70 & 2.74 & 3.97 & 2.80 & 4.01 & 3.73 & 3.53 & 3.76\\
& DriveVA~\citep{liu2026driveva} & \textcolor{red!75!black}{\xmark} & 0.69 & \textbf{1.29} & \textbf{2.03} & \textbf{1.33} & 1.38 & 1.97 & 2.65 & 1.79\\
& UNIVERSE~\citep{liu2026universe} & \textcolor{red!75!black}{\xmark} & \textbf{0.67} & \textbf{1.29} & \textbf{2.03} & \textbf{1.33} & 0.75 & 1.12 & \textbf{1.70} & 1.19\\
\cmidrule(lr){2-11}
& \textbf{DriveWLA (Ours)} & \textcolor{red!75!black}{\xmark} & 0.77 & 1.53 & 2.53 & 1.61 & \textbf{0.34} & \textbf{0.97} & 1.88 & \textbf{1.06}\\
\bottomrule
\end{tabular*}
\end{table*}

\label{sec:grpo}
After imitation learning, we further optimize the Action Expert with Flow-GRPO~\citep{liu2026flow} using driving rewards. During training, SDE noise is added to the flow ODE to sample diverse trajectory candidates for the same scene. These candidates are evaluated using driving rewards, and their relative rewards guide group-relative policy optimization. At inference, the exploration noise is removed, and trajectories are generated using the deterministic flow ODE.

However, the NAVSIM driving reward is multi-dimensional and consists of multiple metrics reflecting different aspects of driving performance. Directly aggregating these metrics into a single scalar reward may obscure individual driving factors and limit effective policy optimization.

To better utilize the optimization signals in driving rewards, we introduce a factorized advantage estimation strategy. Instead of computing the advantage directly from the aggregated PDMS score, we use the individual metric scores to construct a reward vector and normalize each metric within the sampled group:
\begin{equation}
\mathbf r_i=(r_{i,1},...,r_{i,M}),\qquad
z_{i,m}
=
\frac{r_{i,m}-\mu_m}
{\sigma_m+\epsilon},
\end{equation}
where $i$ indexes the sampled trajectory within a group, $m$ indexes the driving metric, and $\mu_m,\sigma_m$ denote the group-wise mean and standard deviation of the corresponding metric. $\epsilon$ is a small constant for numerical stability.

The factorized advantage is computed by weighted aggregation of normalized metric scores, followed by multiplication with a variance-based shrinkage factor:
\begin{equation}
A_i
=
\frac{
\sum_m w_m z_{i,m}
}
{
\sqrt{\operatorname{Var}(\sum_m w_m z_{i,m})+\epsilon}
}
\cdot
\frac{s^2}{s^2+\tau^2},
\label{eq:factorized-advantage}
\end{equation}
where $w_m$ denotes the weight of each driving metric, $s^2$ denotes the variance of PDMS rewards within the sampled group, and $\tau$ controls the shrinkage strength. This formulation preserves the information from individual driving metrics during advantage estimation while improving optimization stability by reducing unreliable updates from low-variance trajectory groups.

Using the proposed advantage during reinforcement learning, we optimize the LoRA~\citep{hu2021lora} parameters of the Action Expert with the Flow-GRPO objective augmented by reference-policy behavioral regularization:
\begin{equation}
\begin{aligned}
\mathcal L ={}&
-\frac{1}{KT}\sum_{i=1}^{K}\sum_{t=1}^{T}
\gamma^{t-1}
\log\pi_{\theta}(a_{t-1}^{i}\mid a_t^{i},\mathcal C_d)A_i\\
&-\frac{\lambda_{\rm BC}}{T}\,
\sum_{t=1}^{T}\mathbb E_{\rm ref}\,
\bigl[\log\pi_{\theta}(\bar a_{t-1}\mid\bar a_t,\mathcal C_d)\bigr].
\end{aligned}
\label{eq:flow-grpo-objective}
\end{equation}
\noindent Here, $K$ is the group size for each scene, $T$ is the number of denoising steps, and $A_i$ is the factorized advantage defined above. The policy $\pi_{\theta}$ specifies stochastic transitions between trajectory states $a_t^i$. With $0<\gamma<1$, later denoising steps receive greater weight, where $t=1$ denotes the final transition from $a_1$ to $a_0$. The expectation $\mathbb E_{\rm ref}$ averages over training conditions $\mathcal C_d$ and fresh reference chains $\bar a=(\bar a_T,\ldots,\bar a_0)$ sampled from the frozen imitation-trained policy $\pi_{\rm ref}(\cdot\mid\mathcal C_d)$. The second term maximizes their transition likelihood under the current policy to retain the imitation-trained behavior, with strength $\lambda_{\rm BC}$. With one PPO inner epoch, the policy-gradient term matches the clipped surrogate's gradient at the rollout parameters, where importance ratios equal one. Sampling and optimization details are provided in the supplementary material.
\par
\endgroup

\section{Experiments}
\label{sec:experiments}
\subsection{Experimental Setup}
\begingroup
\raggedbottom
\widowpenalty=10000
\clubpenalty=10000
\makeatletter\@clubpenalty=10000\makeatother
\noindent\textbf{Datasets and metrics.}
We train DriveWLA on 103,288 NAVSIM~\citep{dauner2024navsim} \texttt{navtrain} scenes and evaluate on 12,146 \texttt{navtest} scenes. 
NAVSIM v1 uses the Predictive Driver Model Score (PDMS), which aggregates no-at-fault collision (NC), drivable-area compliance (DAC), ego progress (EP), time to collision (TTC), and comfort~\citep{dauner2024navsim}. 
NAVSIM v2~\citep{cao2025pseudo} extends PDMS to EPDMS with additional driving-direction compliance (DDC), traffic-light compliance (TLC), lane keeping (LK), history comfort (HC), and extended comfort (EC)~\citep{cao2025pseudo}. 
Detailed metric definitions are provided in the supplementary material.

For zero-shot transfer, we report HD-Score on HUGSIM~\citep{zhou2025hugsim} under closed-loop evaluation, and trajectory L2 error and collision rate on nuScenes~\citep{caesar2020nuscenes} and Bench2Drive~\citep{jia2024bench2drive} under open-loop evaluation.

\noindent\textbf{Implementation details.}
DriveWLA uses a driving-pretrained InternVL3-2B~\citep{zhu2025internvl3,xiong2026recogdrive} as the Language Expert and initializes the World Expert from DriveLaW-Video~\citep{xia2026drivelaw}, which is built on LTX-Video-2B~\citep{hacohen2024ltx}. 
The Action Expert is a 28-block DiT~\citep{peebles2023scalable} that predicts eight $(x,y,\theta)$ waypoints over four seconds conditioned on historical ego poses, ego state, and navigation commands.
During imitation learning, the World DiT, VLM Adapter, and Action DiT are jointly optimized for 30 epochs on 16 A100 GPUs with a batch size of 16 and a learning rate of $3\times10^{-5}$. 
The VLM, video VAE, and text encoder are frozen. 
During reinforcement learning, Flow-GRPO~\citep{liu2026flow} optimizes Action-branch LoRA~\citep{hu2021lora} parameters using driving rewards. 
Unless stated otherwise, inference uses ten Euler steps. Further details are provided in the supplementary material.

\par
\endgroup


\subsection{Main Results}
\noindent\textbf{NAVSIM v1.}
Tab.~\ref{tab:navsim-v1} shows that DriveWLA achieves the best PDMS of 92.3 across approaches from different paradigms. Specifically, it outperforms the best-performing VLM-based and world-model-based planners, SGDrive~\citep{li2026sgdrive} and SimWAM~\citep{zhao2026simwam}, by 1.2 and 0.8 points, respectively, indicating that semantic understanding and future prediction provide complementary benefits for trajectory planning. Furthermore, DriveWLA surpasses existing VLM--world-model integrated planners, including DriveWAM~\citep{shi2026drivewam} and BrainWAM~\citep{zhan2026brainwam}, by 2.2 and 2.8 points, respectively. These improvements are achieved with a more compact expert configuration: DriveWLA uses InternVL3-2B~\citep{zhu2025internvl3} as the Language Expert and LTX-Video-2B~\citep{hacohen2024ltx} as the World Expert, while DriveWAM and BrainWAM adopt 8B/5B and 4B/5B language/video backbones. These results suggest that the advantage of DriveWLA comes from more effective unified World-Language-Action modeling rather than simply scaling up model capacity.

\noindent\textbf{NAVSIM v2.}
Under the corrected official evaluator, DriveWLA achieves the highest EPDMS of 90.1 in Tab.~\ref{tab:navsim-v2}. Importantly, it achieves the best DDC, TTC, LK, and HC scores, while tying for the best TLC. These results show that DriveWLA maintains strong performance across multiple evaluation dimensions, leading to the best overall EPDMS under the more comprehensive NAVSIM v2 evaluation.

\begin{figure*}[!t]
\centering
\includegraphics[width=\linewidth]{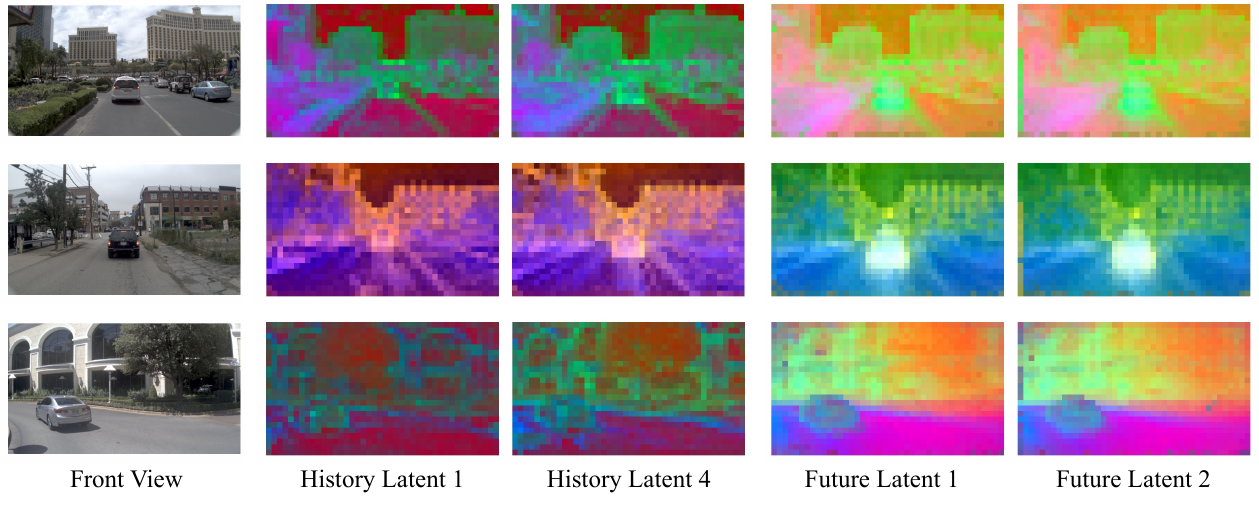}
\caption{\textbf{World Expert latent features.} Each row shows one driving scene. The columns show the front-view image, history latents 1 and 4, and future latents 1 and 2. All PCA-projected latent maps use the same color scale.}
\label{fig:world-representation}
\end{figure*}

\noindent\textbf{Zero-shot on HUGSIM.}
Tab.~\ref{tab:hugsim} shows zero-shot closed-loop evaluation results on the HUGSIM~\citep{zhou2025hugsim} benchmark. DriveWLA achieves the best results on 8 of the 11 reported metrics, including NC, TTC, PDMS, RC, and the average HD-Score. Notably, although the scoring-based DrivoR~\citep{kirby2026driving} performs best on Easy scenarios, its advantage disappears as scenario difficulty increases, whereas DriveWLA achieves higher HD-Scores on Medium, Hard, and Extreme scenarios. These results demonstrate the strong zero-shot generalization and closed-loop driving capability of DriveWLA.

\noindent\textbf{Zero-shot on nuScenes and Bench2Drive.}
On nuScenes, DriveWLA outperforms all zero-shot methods listed in Tab.~\ref{tab:transfer}, achieving the lowest average L2 error of 0.94\,m and collision rate of 0.09\% among them. Compared with fine-tuned planners, DriveWLA achieves a substantially lower collision rate while maintaining a comparable L2 error. On Bench2Drive, DriveWLA continues to demonstrate strong performance, achieving the lowest collision rate of 1.06\% while maintaining competitive L2 performance.

\subsection{Ablation Studies}
\setlength{\textfloatsep}{12pt plus 1pt minus 1pt}
\begingroup
\setlength{\intextsep}{6pt plus 1pt minus 1pt}
\widowpenalty=10000
\clubpenalty=10000
\makeatletter\@clubpenalty=10000\makeatother
We conduct controlled ablations on NAVSIM v1 to assess expert composition, future latents, training strategies, attention masks, and inference-time sampling steps. Unless otherwise noted, each comparison follows the same training and evaluation protocol.

\noindent\textbf{Expert Composition.}
To evaluate the individual contributions of semantic features from the VLM and future dynamic features from the World Expert, we ablate the Language and World Experts separately. As shown in Tab.~\ref{tab:experts}, removing the World Expert reduces PDMS from 90.8 to 86.2, while removing the Language Expert results in a smaller decrease to 89.1. The larger degradation caused by removing the World Expert indicates that predictive dynamics contribute more to planning performance in this setting. Furthermore, the performance degradation after removing each expert demonstrates that semantic and dynamic representations provide complementary benefits, and their joint utilization enables DriveWLA to achieve better performance than using either representation alone.

\begin{table}[!htb]
\centering
\captionsetup{position=top,skip=4pt}
\caption{\textbf{Expert composition.}}
\label{tab:experts}
\footnotesize
\renewcommand{\arraystretch}{1.10}
\setlength{\tabcolsep}{0pt}
\begin{tabular}{@{}>{\raggedright\arraybackslash}p{0.38\linewidth}>{\centering\arraybackslash}p{0.19\linewidth}>{\centering\arraybackslash}p{0.12\linewidth}>{\centering\arraybackslash}p{0.13\linewidth}>{\centering\arraybackslash}p{0.18\linewidth}@{}}
\toprule
\textbf{Experts} & \textbf{Language} & \textbf{World} & \textbf{Action} & \textbf{PDMS$\uparrow$}\\
\midrule
w/o World Expert & \cmark & \xmark & \cmark & 86.2\\
w/o Language Expert & \xmark & \cmark & \cmark & 89.1\\
\rowcolor{scoregraydark}Full DriveWLA & \cmark & \cmark & \cmark & \textbf{90.8}\\
\bottomrule
\end{tabular}
\end{table}

\noindent\textbf{Future-Latent Contribution.}
In the ablation above, removing the World Expert also removes its model capacity and pretrained visual features. To further quantify the contribution of future latents, we first conduct two inference-time ablations using the same imitation-trained checkpoint without retraining the model: setting the future latents to zero or removing them. As shown in Tab.~\ref{tab:future-latents}, these reduce PDMS from 90.8 to 89.7 and 84.1, respectively, confirming that the trained planner indeed uses the future-latent pathway. We further train a history-only variant that retains the same video-pretrained World backbone and parameter capacity but uses only observed-frame latents during both training and inference. Its lower PDMS of 89.8 supports the benefit of future-latent information beyond model capacity and visual pretraining.

\begin{table}[!htb]
\centering
\captionsetup{position=top,skip=4pt}
\caption{\textbf{Contribution of future latents.}}
\label{tab:future-latents}
\label{tab:proposal-future-latents}
\footnotesize
\renewcommand{\arraystretch}{1.10}
\setlength{\tabcolsep}{0pt}
\begin{tabular}{@{}>{\raggedright\arraybackslash}p{0.33\linewidth}>{\centering\arraybackslash}p{0.10\linewidth}>{\centering\arraybackslash}p{0.11\linewidth}>{\centering\arraybackslash}p{0.11\linewidth}>{\centering\arraybackslash}p{0.12\linewidth}>{\centering\arraybackslash}p{0.09\linewidth}>{\columncolor{scoregraylight}[0pt][0pt]\centering\arraybackslash}p{0.14\linewidth}@{}}
\toprule
\textbf{Configuration} & \textbf{NC$\uparrow$} & \textbf{DAC$\uparrow$} & \textbf{TTC$\uparrow$} & \textbf{Comf.$\uparrow$} & \textbf{EP$\uparrow$} & \textbf{PDMS$\uparrow$}\\
\midrule
Full DriveWLA (IL) & 99.2 & 97.7 & 97.2 & 100.0 & 84.0 & \textbf{90.8}\\
Zero future latent & 99.3 & 97.3 & 97.8 & 100.0 & 81.2 & 89.7\\
Remove future latent & 98.2 & 93.6 & 93.2 & 100.0 & 77.7 & 84.1\\
\midrule
History-only retrain & 99.1 & 97.0 & 96.5 & 100.0 & 83.3 & 89.8\\
\bottomrule
\end{tabular}
\end{table}

\noindent\textbf{Training Strategy.}
As shown in Tab.~\ref{tab:world-adaptation}, freezing the pretrained World Expert during imitation learning yields 88.0 PDMS. Jointly optimizing the World Expert raises PDMS from 88.0 to 90.8, highlighting the benefit of adapting its pretrained representations to trajectory planning. Furthermore, applying the original Flow-GRPO after imitation learning improves PDMS to 92.1, with particularly notable improvements in DAC and EP. Flow-GRPO\textsuperscript{$\dagger$} denotes our improved variant with factorized advantage estimation and variance-based shrinkage in Eq.~\eqref{eq:factorized-advantage}, which further improves PDMS to 92.3.
  
\begin{table}[!htb]
\centering
\captionsetup{position=top,skip=4pt}
\caption{\textbf{Training strategy.}}
\label{tab:world-adaptation}
\label{tab:rl}
\footnotesize
\renewcommand{\arraystretch}{1.10}
\setlength{\tabcolsep}{0pt}
\begin{tabular}{@{}>{\raggedright\arraybackslash}p{0.37\linewidth}>{\centering\arraybackslash}p{0.09\linewidth}>{\centering\arraybackslash}p{0.10\linewidth}>{\centering\arraybackslash}p{0.10\linewidth}>{\centering\arraybackslash}p{0.11\linewidth}>{\centering\arraybackslash}p{0.09\linewidth}>{\centering\arraybackslash}p{0.14\linewidth}@{}}
\toprule
\textbf{Configuration} & \textbf{NC$\uparrow$} & \textbf{DAC$\uparrow$} & \textbf{TTC$\uparrow$} & \textbf{Comf.$\uparrow$} & \textbf{EP$\uparrow$} & \textbf{PDMS$\uparrow$}\\
\midrule
Frozen World Expert & 98.5 & 96.1 & 96.4 & \textbf{100.0} & 80.4 & 88.0\\
Trainable World Expert & \textbf{99.2} & 97.7 & \textbf{97.2} & \textbf{100.0} & 84.0 & 90.8\\
$+$ Flow-GRPO & 99.1 & 98.7 & 97.1 & \textbf{100.0} & 86.1 & 92.1\\
\rowcolor{scoregraydark}$+$ Flow-GRPO\textsuperscript{$\dagger$} & 99.1 & \textbf{98.8} & 97.0 & \textbf{100.0} & \textbf{86.4} & \textbf{92.3}\\
\bottomrule
\end{tabular}
\end{table}

\begin{table}[!htb]
\centering
\captionsetup{position=top,skip=4pt}
\caption{\textbf{Cross-Expert information flow.}}
\label{tab:masks}
\footnotesize
\renewcommand{\arraystretch}{1.10}
\setlength{\tabcolsep}{0pt}
\begin{tabular}{@{}>{\centering\arraybackslash}p{0.18\linewidth}>{\centering\arraybackslash}p{0.18\linewidth}>{\centering\arraybackslash}p{0.22\linewidth}>{\centering\arraybackslash}p{0.22\linewidth}>{\centering\arraybackslash}p{0.20\linewidth}@{}}
\toprule
\textbf{W $\rightarrow$ L} &
\textbf{L $\rightarrow$ W} &
\textbf{W / L $\rightarrow$ A} &
\textbf{A $\rightarrow$ W / L} &
\textbf{PDMS$\uparrow$}\\
\midrule
\rowcolor{scoregraydark}\cmark & \cmark & \cmark & \xmark & \textbf{90.8}\\
\cmark & \xmark & \cmark & \xmark & 90.1\\
\xmark & \cmark & \cmark & \xmark & 89.8\\
\cmark & \cmark & \cmark & \cmark & 90.4\\
\xmark & \xmark & \cmark & \xmark & 88.4\\
\bottomrule
\end{tabular}
\end{table}

\noindent\textbf{Cross-Expert Information Flow.}
To explore proper information flow among the three experts in our Cross-Expert Attention module, we design different attention masks to control the visibility between the World, Language, and Action Experts. Here, $W$, $L$, and $A$ denote the World, Language, and Action Experts, respectively, and $X \rightarrow Y$ indicates information flow from Expert $X$ to Expert $Y$. As shown in Tab.~\ref{tab:masks}, the bidirectional World--Language interaction with the Action Expert only receiving representations from both experts achieves the highest PDMS of 90.8. Removing either direction of World--Language information flow degrades performance, while removing both directions causes a further drop. These results demonstrate that bidirectional information exchange between semantic and dynamic representations is beneficial for leveraging their complementary capabilities. Moreover, allowing information to flow from the Action Expert to the World and Language Experts leads to performance degradation, indicating that introducing action representations into the World and Language Experts is not beneficial for planning.

\begin{table}[!htb]
\centering
\captionsetup{position=top,skip=4pt}
\caption{\textbf{Sampling steps.}}
\label{tab:steps}
\footnotesize
\renewcommand{\arraystretch}{1.10}
\setlength{\tabcolsep}{0pt}
\begin{tabular}{@{}>{\centering\arraybackslash}p{0.10\linewidth}>{\centering\arraybackslash}p{0.11\linewidth}>{\centering\arraybackslash}p{0.12\linewidth}>{\centering\arraybackslash}p{0.12\linewidth}>{\centering\arraybackslash}p{0.14\linewidth}>{\centering\arraybackslash}p{0.10\linewidth}>{\centering\arraybackslash}p{0.14\linewidth}>{\centering\arraybackslash}p{0.17\linewidth}@{}}
\toprule
\textbf{Steps} & \textbf{NC$\uparrow$} & \textbf{DAC$\uparrow$} & \textbf{TTC$\uparrow$} & \textbf{Comf.$\uparrow$} & \textbf{EP$\uparrow$} & \textbf{PDMS$\uparrow$} & \textbf{Latency$\downarrow$}\\
\midrule
1 & 91.2 & 81.4 & 63.6 & 1.7 & 68.3 & 49.2 & \textbf{540} ms\\
5 & 99.2 & 97.6 & 96.8 & \textbf{100.0} & \textbf{84.1} & 90.6 & 810 ms\\
\rowcolor{scoregraydark}10 & 99.2 & 97.7 & 97.2 & \textbf{100.0} & 84.0 & \textbf{90.8} & 970 ms\\
15 & \textbf{99.3} & \textbf{97.8} & 97.5 & \textbf{100.0} & 83.4 & \textbf{90.8} & 1160 ms\\
20 & \textbf{99.3} & \textbf{97.8} & \textbf{97.6} & \textbf{100.0} & 83.3 & 90.7 & 1390 ms\\
\bottomrule
\end{tabular}
\end{table}

\noindent\textbf{Sampling Steps.}
With a single Euler step, DriveWLA achieves only 49.2 PDMS (Tab.~\ref{tab:steps}). Increasing the sampling steps to five recovers
most of the performance, achieving 90.6 PDMS, while ten steps further improve PDMS to the best
result of 90.8 at 970\,ms latency. Increasing the steps beyond ten provides no additional improvement at 15 steps and causes a slight performance drop at 20 steps, while \mbox{introducing} higher inference latency. We therefore adopt \mbox{ten steps}, where planning performance reaches its highest measured level before additional sampling increases latency without improving PDMS.

\par\endgroup

\subsection{Representation Visualization}

Fig.~\ref{fig:world-representation} shows PCA projections of the World Expert's history and future features after imitation learning in three driving scenes. The four observed history frames are encoded separately, yielding four history latents; history latents 1 and 4 therefore correspond to the first and fourth observed frames. The two future latents are initialized from Gaussian noise and generated in a single forward pass by the World Expert conditioned on past and current images. The two future latents represent the first and second 2-second intervals of the 4-second prediction horizon, respectively.

Importantly, the Future Latent maps represent the World Expert's predictions of the scene beyond the observed frames. Their changing patterns from the first to the second 2-second interval suggest that the model predicts how the scene will evolve over time, including the motion of objects and changes in their positions. This provides qualitative evidence that the unified architecture captures not only the observed scene but also its future evolution, supplying the Action Expert with useful predictive context for generating trajectories in dynamic scenes.

\section{Conclusion}
We proposed DriveWLA, a unified World-Language-Action framework for autonomous driving. Our results show that semantic understanding and future dynamics provide complementary information for trajectory planning, and the controlled cross-expert interaction can effectively integrate these capabilities. DriveWLA achieves state-of-the-art performance on NAVSIM and strong zero-shot generalization across open-loop and closed-loop benchmarks. We hope DriveWLA can serve as a baseline and that World-Language-Action modeling can become a promising paradigm for future autonomous driving research.

\label{main-text-end}
\clearpage
\subsection*{LLM Usage Statement}
We used DeepSeek to assist with drafting and editing the manuscript to improve clarity and readability, as well as to identify relevant prior work. Additionally, we utilized DeepSeek Harness to assist with code development, implementation, debugging, and refinement. All AI-assisted content and code were reviewed, verified, and validated, and we take full responsibility for the final manuscript and implementation.

{
    \small
    \bibliographystyle{ieeenat_fullname}
    \bibliography{main}
}

\maketitlesupplementary
\appendix
\setcounter{equation}{0}
\setcounter{table}{0}
\setcounter{figure}{0}
\renewcommand{\theequation}{S\arabic{equation}}
\renewcommand{\thetable}{S\arabic{table}}
\renewcommand{\thefigure}{S\arabic{figure}}
\renewcommand{\theHequation}{S\arabic{equation}}
\renewcommand{\theHtable}{S\arabic{table}}
\renewcommand{\theHfigure}{S\arabic{figure}}
\section{Implementation Details}
\label{app:implementation}

\subsection{Expert Configuration}

\noindent\textbf{World Expert.}
The World Expert is initialized from DriveLaW-Video~\citep{xia2026drivelaw}, which is built on LTX-Video-2B~\citep{hacohen2024ltx}. Its video VAE encodes four observed front-view frames at $768\times1344$ resolution. A 28-block DiT with hidden width 2048 processes these latents together with two Gaussian-initialized future latents. The video VAE and text encoder remain frozen throughout training. The World DiT is optimized during imitation learning and frozen during Flow-GRPO~\citep{liu2026flow} refinement.

\noindent\textbf{Language Expert.}
The Language Expert uses the driving-pretrained InternVL3-2B~\citep{zhu2025internvl3} checkpoint from ReCogDrive~\citep{xiong2026recogdrive}. The frozen VLM takes the current front-view image with dynamic $448\times448$ tiling and the driving prompt $P$ in Eq.~(3) of the main paper. Its final-layer hidden states $H_L$ for valid input tokens are passed to a 28-block VLM Adapter with hidden width 512. The adapter is trained during imitation learning and frozen during Flow-GRPO refinement.

\noindent\textbf{Action Expert.}
The Action Expert is a 28-block DiT with hidden width 512. The driving condition $\mathcal C_d$ consists of four historical ego poses sampled at 2\,Hz, the current ego state, and the navigation command. The Action Expert predicts eight $(x,y,\theta)$ waypoints at 0.5\,s intervals over four seconds in the current ego-centric frame. We optimize the full DiT during imitation learning and update only the Action-branch LoRA parameters during Flow-GRPO refinement.

During imitation learning, we jointly train the World DiT, VLM Adapter, and Action DiT for 30 epochs using the flow-matching objective in Eq.~(4) of the main paper. We use 16 A100 GPUs, a global batch size of 16, and an initial learning rate of $3\times10^{-5}$ with cosine decay. Expert trajectories provide the only supervision, and the trajectory loss updates the World DiT and VLM Adapter through Cross-Expert Attention.

\subsection{Trajectory Sampling}

At inference, the World Expert processes the observed latents $Z_{\rm obs}$ and Gaussian-initialized future latents $Z_{\rm fut}$ in one forward pass. The Action Expert denoises a trajectory using World and Language features from Cross-Expert Attention, together with ego history, state, and navigation commands $\mathcal C_d$. For simplicity, we do not explicitly write the World and Language features as conditioning inputs in the equations. The Action Expert starts from $a_{\tau_0}=a_1\sim\mathcal N(\mathbf 0,\mathbf I)$ and follows the learned velocity field from $\tau_0=1$ to $\tau_T=0$. By default, it uses $T=10$ Euler steps:
\begin{equation}
\begin{gathered}
a_{\tau_{k+1}}=a_{\tau_k}+(\tau_{k+1}-\tau_k)
v_{\theta}(a_{\tau_k},\tau_k,\mathcal C_d),
\\
k=0,\ldots,T-1.
\end{gathered}
\label{eq:app-euler}
\end{equation}

\subsection{Text Conditioning}
\label{app:text-conditioning}

\noindent\textbf{World Expert.}
The World Expert uses a fixed video-generation caption for all scenes during both training and inference.
\begin{quote}
\small
A high-quality, photorealistic dashboard camera view of autonomous driving. Based on the past 2 seconds videos, predict and generate the next 4 seconds of realistic driving continuation, Maintain temporal consistency, stable camera perspective, natural motion flow without jitter or artifacts, clear details, and realistic physics.
\end{quote}
\noindent\textbf{Language Expert.}
The prompt $P$ contains four historical ego poses, the navigation command (\texttt{TURN LEFT}, \texttt{GO STRAIGHT}, or \texttt{TURN RIGHT}), and instructions for the output format of eight $(x,y,\theta)$ waypoints. 

\section{Evaluation Protocols}
\label{app:datasets}
\label{app:metrics}

\subsection{NAVSIM v1}

On NAVSIM v1~\citep{dauner2024navsim}, DriveWLA is trained on 103,288 \texttt{navtrain} scenes and evaluated on 12,146 \texttt{navtest} scenes. The Predictive Driver Model Score (PDMS) applies multiplicative penalties for no-at-fault collisions (NC) and drivable-area compliance (DAC) to a weighted combination of ego progress (EP), time-to-collision compliance (TTC), and comfort (C, shown as Comf.\ in the tables):
\begin{equation}
\mathrm{PDMS}=\mathrm{NC}\times\mathrm{DAC}\times
\left(\frac{5\,\mathrm{EP}+5\,\mathrm{TTC}+2\,\mathrm{C}}{12}\right).
\label{eq:app-pdms}
\end{equation}
NC takes values in $\{0,0.5,1\}$ depending on the type of at-fault collision. DAC checks whether the simulated ego footprint stays in the drivable area. EP measures route progress relative to the reference planner, TTC checks for collisions within a fixed time bound, and C checks motion comfort. The PDMS and EPDMS formulas use $[0,1]$ scores; the tables report these metrics on a $[0,100]$ scale.

The ResNet-34 listed for DriveWorld-VLA in Tab.~1 of the main paper is its BEV encoder for latent world modeling~\citep{liu2026driveworld}.

\subsection{NAVSIM v2}

Tab.~2 of the main paper reports results on the full \texttt{navtest} split using the corrected official NAVSIM v2 evaluator with human-reference filtering~\citep{cao2025pseudo}. EPDMS adds driving-direction compliance (DDC) and traffic-light compliance (TLC) as multiplicative penalties, and includes lane keeping (LK), history comfort (HC), and extended comfort (EC) in the weighted score:
\begin{equation}
\begin{gathered}
\mathrm{EPDMS}=\mathrm{NC}\times\mathrm{DAC}\times\mathrm{DDC}\times\mathrm{TLC}\\
\times\left(\frac{5\,\mathrm{EP}+5\,\mathrm{TTC}+2\,\mathrm{LK}+2\,\mathrm{HC}+2\,\mathrm{EC}}{16}\right).
\end{gathered}
\label{eq:app-epdms}
\end{equation}
DDC penalizes driving against the lane direction, TLC penalizes red-light violations, and LK checks lane adherence. HC measures how smoothly the plan continues the observed ego motion. Before aggregation, the human-reference filter sets NC, DAC, DDC, TLC, EP, TTC, LK, or HC to one if the human scores zero on that metric.

EC measures consistency between overlapping parts of successive plans.

\subsection{HUGSIM}

We evaluate DriveWLA zero-shot in closed loop on all 436 HUGSIM~\citep{zhou2025hugsim} scenarios, which include 88 from nuScenes~\citep{caesar2020nuscenes}, 113 from KITTI-360~\citep{liao2022kitti}, 108 from Waymo~\citep{sun2020scalability}, and 127 from PandaSet~\citep{xiao2021pandaset}. For the main comparison, we follow WA-JEPA's evaluation setup~\citep{wang2026wa} and run the simulator synchronously (Sec.~\ref{app:closed-loop}).

At each simulation step, HUGSIM computes $\mathrm{NC}\times\mathrm{DAC}\times(5\,\mathrm{TTC}+2\,\mathrm{C})/7$. The rollout average gives PDMS, and the scenario's HD-Score is PDMS multiplied by route completion (RC). NC, DAC, TTC, C, PDMS, and RC in Tab.~3 of the main paper are averaged over all 436 scenarios.

The Easy, Medium, Hard, and Extreme levels contain 80, 157, 96, and 103 scenarios. Within each level, we average scenes within each source dataset and then average the four datasets equally. The overall HD-Score weights the four levels by their scenario counts~\citep{wang2026wa}.

We take the VAD, UniAD, LTF, and DrivoR results in Tab.~3 from the WA-JEPA reevaluation~\citep{wang2026wa} and report them on a $[0,100]$ scale. We use the same scenarios, driving commands, controller, metrics, and aggregation as WA-JEPA. DriveWLA uses only a front-view camera; LTF uses three cameras, while VAD, UniAD, and DrivoR use four.

\subsection{nuScenes}

We evaluate DriveWLA zero-shot on the full 150-scene nuScenes~\citep{caesar2020nuscenes} validation split. Following UniAD~\citep{hu2023planning}, we report L2 error between the predicted and logged ego positions and collision rate based on overlap between the predicted ego footprint and annotated objects. We report both metrics at 1, 2, and 3 seconds and their averages.

Tab.~4 uses the reported UniAD~\citep{hu2023planning}, OpenDriveVLA~\citep{zhou2026opendrivevla}, and UniDriveVLA-L~\citep{li2026unidrivevla} results under the same metrics. The DriveVLA-W0~\citep{li2026drivevla} and PWM~\citep{zhao2026forecasting} zero-shot results are taken from DriveVA~\citep{liu2026driveva}.

\subsection{Bench2Drive}

On Bench2Drive~\citep{jia2024bench2drive}, we evaluate DriveWLA zero-shot on recorded observations using the open-loop trajectory protocol in DriveVA~\citep{liu2026driveva}. We report L2 error and collision rate at 1, 2, and 3 seconds and their averages.

All methods in the Bench2Drive comparison are trained on NAVSIM and evaluated zero-shot. The DriveVLA-W0, PWM, and DriveVA results are taken from DriveVA~\citep{liu2026driveva}, and the UNIVERSE results from its original report~\citep{liu2026universe}.

\begin{table*}[!t]
\centering
\caption{\textbf{Future-motion prediction from future features.} These features are extracted at future-latent positions. IL denotes imitation learning. Errors are in meters. Bold values denote the lowest error within each probe.}
\label{tab:supp-motion-probes}
\begingroup
\small
\setlength{\tabcolsep}{3pt}
\renewcommand{\arraystretch}{1.15}
\begin{tabular*}{\textwidth}{@{\extracolsep{\fill}}lcccccc@{}}
\toprule
\multirow{2}{*}{\textbf{Input / configuration}} & \multicolumn{3}{c}{\textbf{All objects}} & \multicolumn{3}{c}{\textbf{Moving objects}}\\
\cmidrule(lr){2-4}\cmidrule(l){5-7}
& ADE@2s & ADE@4s & FDE@4s & ADE@2s & ADE@4s & FDE@4s\\
\midrule
\multicolumn{7}{l}{\emph{Predicting Motion from Future Features}}\\
Position only & 2.345 & 4.034 & 7.062 & 5.013 & 8.718 & 15.798\\
Position + future (before IL) & 1.963 & 3.408 & 6.009 & 3.789 & 6.619 & 11.983\\
Position + future (after IL) & \textbf{1.909} & \textbf{3.285} & \textbf{5.745} & \textbf{3.458} & \textbf{6.013} & \textbf{10.800}\\
\midrule
\multicolumn{7}{l}{\emph{Improving History-Based Prediction with Future Features}}\\
Predictor only & 0.457 & 1.066 & 2.402 & 0.715 & 1.795 & 4.310\\
Predictor + MLP (zero future features) & 0.433 & 1.026 & 2.332 & 0.698 & 1.770 & 4.268\\
Predictor + MLP (future features before IL) & 0.429 & 0.997 & 2.243 & 0.666 & 1.639 & 3.901\\
Predictor + MLP (future features after IL) & \textbf{0.425} & \textbf{0.986} & \textbf{2.217} & \textbf{0.655} & \textbf{1.596} & \textbf{3.776}\\
\bottomrule
\end{tabular*}
\endgroup
\end{table*}

\begin{figure*}[!t]
\centering
\begin{minipage}{\textwidth}
\centering
\includegraphics[width=\linewidth]{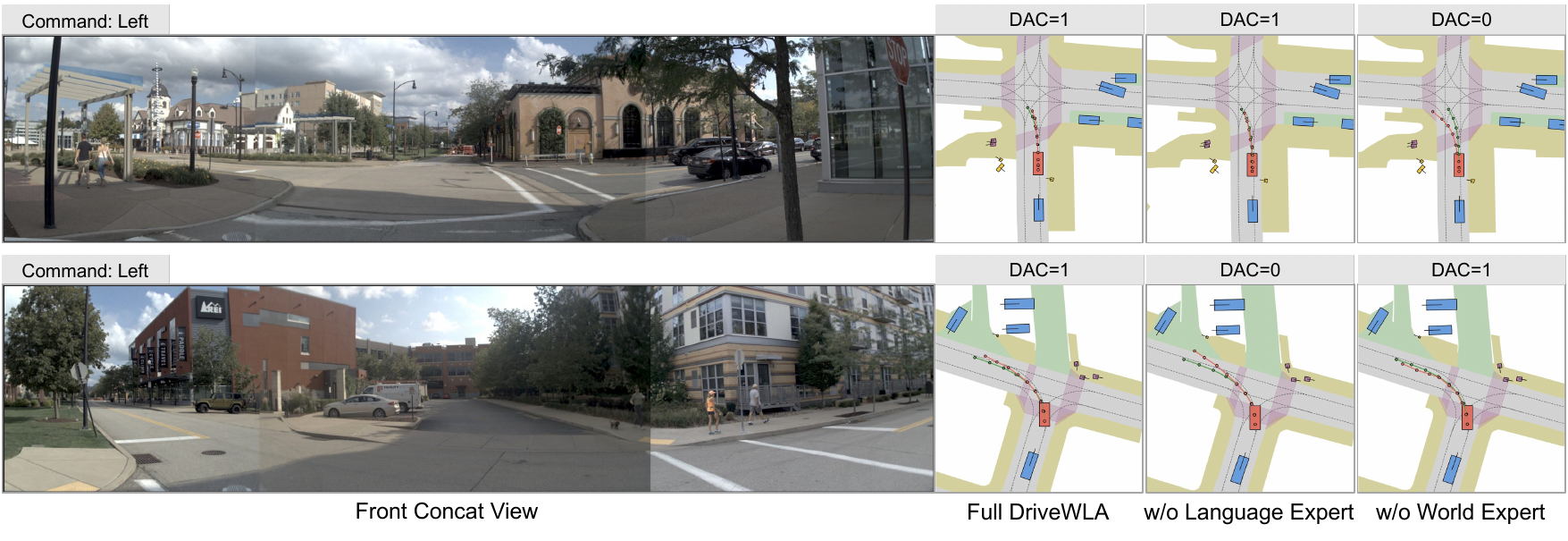}
\par\smallskip
{\footnotesize (a) Expert ablations on NAVSIM \texttt{navtest}.\par}
\end{minipage}
\par\medskip
\begin{minipage}{\textwidth}
\centering
\includegraphics[width=\linewidth]{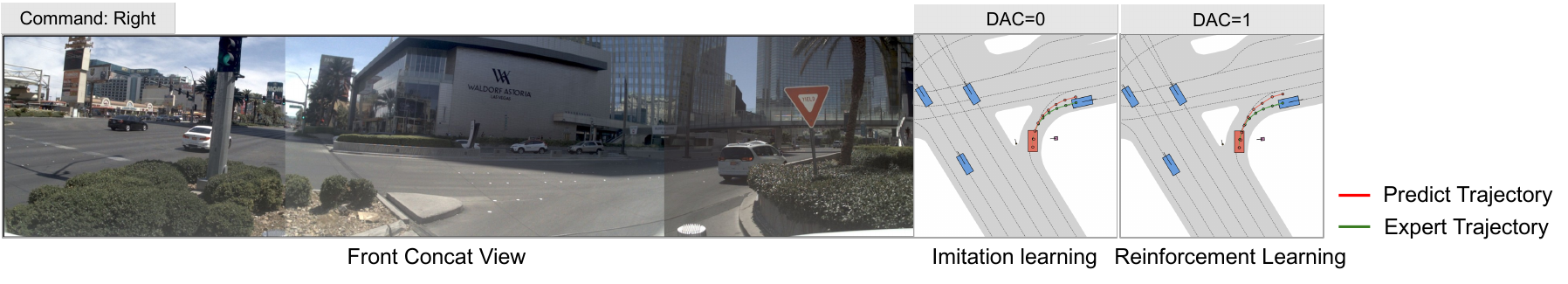}
\par\smallskip
{\footnotesize (b) Before and after RL refinement on NAVSIM \texttt{navtest}.\par}
\end{minipage}
\caption{\textbf{Expert ablations and RL refinement.}
(a) Removing the World or Language Expert causes the vehicle to leave the drivable area in different left-turn scenes.
(b) After RL refinement, the policy stays within the drivable area while taking a different path from the expert demonstration.}
\label{fig:supp-qualitative}
\end{figure*}

\begin{figure*}[!t]
\centering
\includegraphics[width=0.92\textwidth]{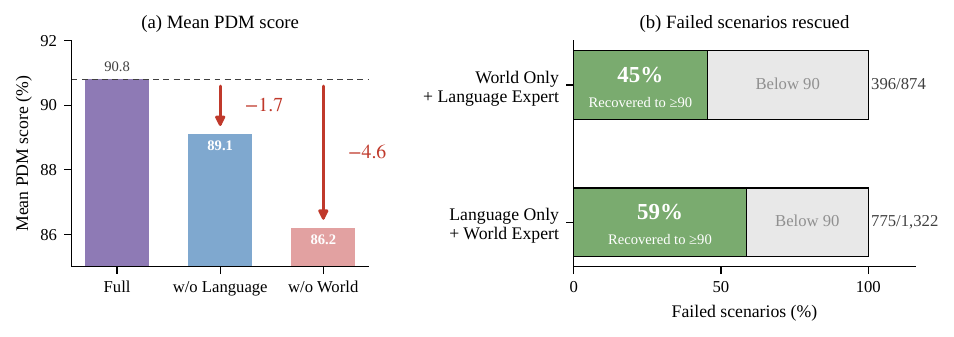}
\caption{\textbf{Expert ablation on NAVSIM \texttt{navtest}.}
(a) Average PDMS over 12,146 scenes. Dashed line marks the full model's score; arrows show drops after expert removal.
(b) We select scenes where the World-only or Language-only model scores PDM $\leq0.6$ and evaluate Full DriveWLA on the same scenes. Green shows the percentage with per-scene PDM $\geq0.9$ (90 in the plot); gray shows the rest.}
\label{fig:supp-branch-statistics}
\end{figure*}

\begin{table}[!t]
\centering
\caption{\textbf{Effect of the driving prompt on the World Expert.} Caption-only baselines are from Tab.~5 of the main paper.}
\label{tab:supp-world-prompts}
\begingroup
\footnotesize
\setlength{\tabcolsep}{2pt}
\renewcommand{\arraystretch}{1.15}
\begin{tabular*}{\linewidth}{@{\extracolsep{\fill}}cl>{\columncolor{scoregraylight}[\tabcolsep][0pt]}c@{}}
\toprule
\textbf{Language Expert} & \textbf{World text} & \textbf{PDMS}$\uparrow$\\
\midrule
\cmark & Caption & 90.8\\
\xmark & Caption & 89.1\\
\xmark & Caption + driving prompt & 89.7\\
\cmark & Caption + driving prompt & 90.8\\
\bottomrule
\end{tabular*}
\endgroup
\end{table}

\begin{table*}[!t]
\centering
\caption{\textbf{Effect of synchronous and asynchronous execution on HUGSIM.} All evaluations use the full set of 436 HUGSIM scenarios. Higher is better for all reported metrics.}
\label{tab:supp-execution}
\begingroup
\footnotesize
\setlength{\tabcolsep}{2pt}
\renewcommand{\arraystretch}{1.0}
\begin{tabular*}{\textwidth}{@{\extracolsep{\fill}}lccccccccccc>{\columncolor{scoregraylight}[\tabcolsep][0pt]}c@{}}
\toprule
\multirow{2}{*}{\textbf{Execution}} & \multirow{2}{*}{\textbf{Steps}} & \multicolumn{6}{c}{\textbf{All 436 scenarios}} & \multicolumn{5}{c}{\textbf{HD-Score}}\\
\cmidrule(lr){3-8}\cmidrule(l){9-13}
& & NC & DAC & TTC & Comf. & PDMS & RC & Easy & Medium & Hard & Extreme & Overall\\
\midrule
Synchronous & 5 & 67.47 & 93.78 & 58.47 & 48.95 & 49.42 & 57.25 & 64.89 & 44.81 & 28.05 & 17.65 & 38.39\\
Synchronous & 10 & 65.80 & 95.06 & 59.27 & 75.78 & 54.92 & 54.99 & 73.87 & 47.91 & 31.41 & 15.87 & 41.47\\
\midrule
Asynchronous & 5 & 72.72 & 93.65 & 61.34 & 18.81 & 45.60 & 41.29 & 53.30 & 33.28 & 13.64 & 6.54 & 26.31\\
Asynchronous & 10 & 73.30 & 93.26 & 63.48 & 41.20 & 50.65 & 42.70 & 60.00 & 36.97 & 14.45 & 8.03 & 29.40\\
\bottomrule
\end{tabular*}
\endgroup
\end{table*}

\section{Flow-GRPO Training Details}
\label{app:policy-refinement}

For policy refinement, we select training scenes where the imitation policy scores below 90 PDMS on the $[0,100]$ scale.
For each scene, we sample $K=8$ trajectories using Flow-GRPO-style score-corrected stochastic transitions~\citep{liu2026flow} on the linear-quadratic time grid from the LTX-Video pipeline. Each chain runs from Gaussian noise $a_T^i$ to a sampled trajectory $a_0^i$. The subscripts label the discrete denoising states. The starting state $a_T^i$ is at flow time $\tau=1$, corresponding to $a_1$ in Sec.~3.1 of the main paper. We score the final trajectories to compute $A_i$. The same advantage is used for every transition in chain $i$ in Eq.~(9).

We standardize NC, DAC, EP, TTC, and C within each group and compute the factorized advantage in Eq.~(8), with $w_{\rm NC}=w_{\rm DAC}=w_{\rm C}=2$ and $w_{\rm EP}=w_{\rm TTC}=5$.
The NAVSIM scorer returns PDMS on a $[0,1]$ scale. We use the population variance of these raw scores for $s^2$ and set the shrinkage parameter $\tau=0.1$. This fixed parameter is distinct from the sampler's flow time $\tau_k$.

Training runs for eight epochs with $T=10$ denoising steps per chain, discount $\gamma=0.6$, and reference-policy coefficient $\lambda_{\rm BC}=0.1$. Here, $T$ counts denoising steps; in the problem definition in Sec.~3 of the main paper, it counts future trajectory steps. Only the Action Expert LoRA parameters are updated, with rank 16 and scaling factor 32. At inference, we use the Euler sampler in Eq.~\eqref{eq:app-euler}.

For the regularization term in Eq.~(9), we sample fresh chains $\bar a=(\bar a_T,\ldots,\bar a_0)$ from the frozen imitation-trained policy $\pi_{\rm ref}(\cdot\mid\mathcal C_d)$. We minimize their average negative transition log likelihood under the current policy, using the same driving conditions $\mathcal C_d$.

The index $t$ runs from $T$ to 1, so $\gamma^{t-1}$ gives the greatest weight to the final transition from $a_1$ to $a_0$. We use one PPO inner epoch. At the rollout parameters, the importance ratio is one and PPO ratio clipping is inactive. The policy-gradient term in Eq.~(9) therefore has the same gradient as the clipped surrogate at that point, before the numerical processing below.

For numerical stability, standard-normal noise sampled at each stochastic transition is clipped to $[-5,5]$, and the transition standard deviation is bounded below by $10^{-3}$. We clip elementwise Gaussian log densities to $[-5,2]$ and average them over waypoint and action dimensions when computing the optimization scores.

Tabs.~5, 6, 8, and 9 of the main paper evaluate expert composition, future latents, information flow, and sampling steps, respectively, using imitation-trained models. Tab.~7 reports 90.8 PDMS for the full IL model, 92.1 for Flow-GRPO with aggregate PDMS rewards, and 92.3 for our factorized variant, Flow-GRPO$^{\dagger}$.

\section{World Expert Analysis}
\label{app:world-analysis}

\subsection{Future-Latent Ablations}
\label{app:future-ablations}

We perform two inference-time ablations and include a separately trained history-only model (Tab.~6 of the main paper). All models are evaluated before RL refinement with ten Action Euler steps.

\noindent\textbf{Inference-time interventions.}
Both ablations use the IL checkpoint. One sets the features at the future-latent positions to zero before every Cross-Expert Attention block. Token positions and the mask stay unchanged. The other removes them and uses only the four observed-frame latents.

On the full \texttt{navtest} split, zeroing and removing future latents reduce PDMS from 90.8 to 89.7 and 84.1, respectively. The drop after zeroing shows that the planner benefits from the future latents. Removing those causes a larger drop, though this also changes the attention structure.

\noindent\textbf{History-only training.}
To assess a history-only model without removing future tokens from the full model at inference, we train a separate model using only observed-frame latents. It keeps the same video-pretrained backbone, parameter capacity, Language and Action Experts, and trajectory objective. It uses only the four observed-frame latents during training and inference. Its PDMS is 89.8, compared with 90.8 for the full model, with lower DAC, TTC, and EP. 

\subsection{Future-Motion Probes}

We train two MLP probes to predict each surrounding object's cumulative 2D displacement from its current position at eight future times, spaced 0.5\,s apart over four seconds. Targets, predictions, and ADE/FDE all use the ego frame fixed at the current timestep. We extract future-latent features before the final Cross-Expert Attention block with DriveWLA frozen; only the World weights differ between pretrained and IL configurations.

We train the probes on \texttt{navtrain} and evaluate them on \texttt{navtest}. Moving objects are defined as those with an observed speed above 1\,m/s. ADE@2s and ADE@4s measure average displacement error over the first two and four seconds, respectively. FDE@4s measures displacement error at four seconds.


\noindent\textbf{Probe training.} For each object, we extract features from both future latents within its current image box (ROI) and over the full image (global). We concatenate these features, reduce them to 128 dimensions with PCA, and concatenate the result with the object's current 2D position. PCA is fitted separately for each feature configuration using training data only. Input normalization also uses training data only. Both probes use a 64-unit hidden layer with GELU. We train them for 30 epochs with masked Smooth $L_1$ loss, AdamW, a learning rate of $10^{-3}$, weight decay of $10^{-4}$, and batches of 512 objects.

\noindent\textbf{Predicting Motion from Future Features.}
The position-only baseline receives only the object's current position. On moving objects, adding future features before IL reduces ADE@4s/FDE@4s from 8.718/15.798\,m to 6.619/11.983\,m; after IL, the errors fall further to 6.013/10.800\,m (Tab.~S1).

\noindent\textbf{Improving History-Based Prediction with Future Features.}
We next test whether future features provide additional predictive information beyond motion history. We freeze a motion predictor trained on a 14-dimensional observed state: four 2D object positions, their validity flags, and velocity estimated from the history. An MLP adds a correction to the frozen prediction, using the current position and future features. For comparison, we train the same MLP with future features set to zero. Compared with this control, adding future features after IL reduces moving-object ADE@4s from 1.770 to 1.596\,m (9.8\%) and FDE@4s from 4.268 to 3.776\,m (11.5\%).

\subsection{Interpreting the Latent Visualizations}
Fig.~4 of the main paper shows World features after imitation learning. The history maps correspond to the first and fourth observed frames; the future maps correspond to the first and second 2-second intervals of the 4-second horizon. These future features are inferred in one forward pass from observed images and Gaussian-initialized latents, rather than encoded from ground-truth future frames. The motion probes above measure how accurately future object displacements can be predicted from these features. 

\section{Language Expert Analysis}
\label{app:language-analysis}

We test the fixed World caption alone and with the driving prompt added. Each condition is evaluated with and without the Language Expert.

Without the Language Expert, adding the prompt raises PDMS from 89.1 to 89.7 (Tab.~\ref{tab:supp-world-prompts}), below the full model's 90.8. With the Language Expert, both prompt settings score 90.8. Providing the driving prompt to the World Expert therefore does not fully replace the Language Expert.

\section{Closed-Loop Execution}
\label{app:closed-loop}

\noindent\textbf{Execution protocols.}
Tab.~3 of the main paper uses ten Action Euler steps and synchronous execution: the simulator pauses during inference. In asynchronous execution, it keeps advancing and rendering frames, and applies each trajectory when inference finishes. We compare both protocols on the same 436 scenarios with five and ten Action Euler steps.

\noindent\textbf{Computation time.}
The times in Tab.~9 of the main paper are measured on an NVIDIA A100 and include the VLM forward pass, trajectory inference, and action decoding. In HUGSIM, adding 3D Gaussian splatting rendering and image preprocessing gives mean end-to-end times of 1.61\,s with five steps and 1.73\,s with ten on the same GPU.

\noindent\textbf{Results.}
Asynchronous execution reduces HD-Score from 38.39 to 26.31 with five steps and from 41.47 to 29.40 with ten (Tab.~\ref{tab:supp-execution}). Route completion and comfort also decrease, while NC and TTC increase. Ten steps improve PDMS and HD-Score under both protocols. Delayed trajectory updates remain a limitation in closed-loop execution.

\section{Qualitative Analysis}
\label{app:qualitative}

We compare trajectories on NAVSIM \texttt{navtest}.

\noindent\textbf{Expert complementarity.}
In the first left-turn scene in Fig.~\ref{fig:supp-qualitative}(a), the model without the World Expert leaves the drivable area, while the model without the Language Expert succeeds. In the second scene, the model without the Language Expert leaves the drivable area. Full DriveWLA stays within the drivable area in both scenes.

\noindent\textbf{RL refinement.}
In Fig.~\ref{fig:supp-qualitative}(b), the imitation policy leaves the drivable area. RL refinement raises DAC from 0 to 1 while producing a trajectory different from the expert demonstration.

\noindent\textbf{Expert ablation across scenes.}
The three models in Fig.~\ref{fig:supp-branch-statistics} are trained separately with imitation learning and use ten Action Euler steps. Across 12,146 scenes, removing the Language and World Experts lowers mean PDMS from 90.8 to 89.1 and 86.2, respectively (Tab.~5 of the main paper). We select failures with PDM $\leq0.6$ from each single-expert model, then evaluate Full DriveWLA on those same scenes. Full DriveWLA reaches per-scene PDM $\geq0.9$ in 396/874 World-only failures (45\%) and 775/1,322 Language-only failures (59\%). The results show experts complement each other: adding the missing expert often resolves failures of either single-expert model, though some scenes remain challenging.

\begin{figure*}[p]
\centering
\includegraphics[width=\textwidth,height=0.89\textheight,keepaspectratio]{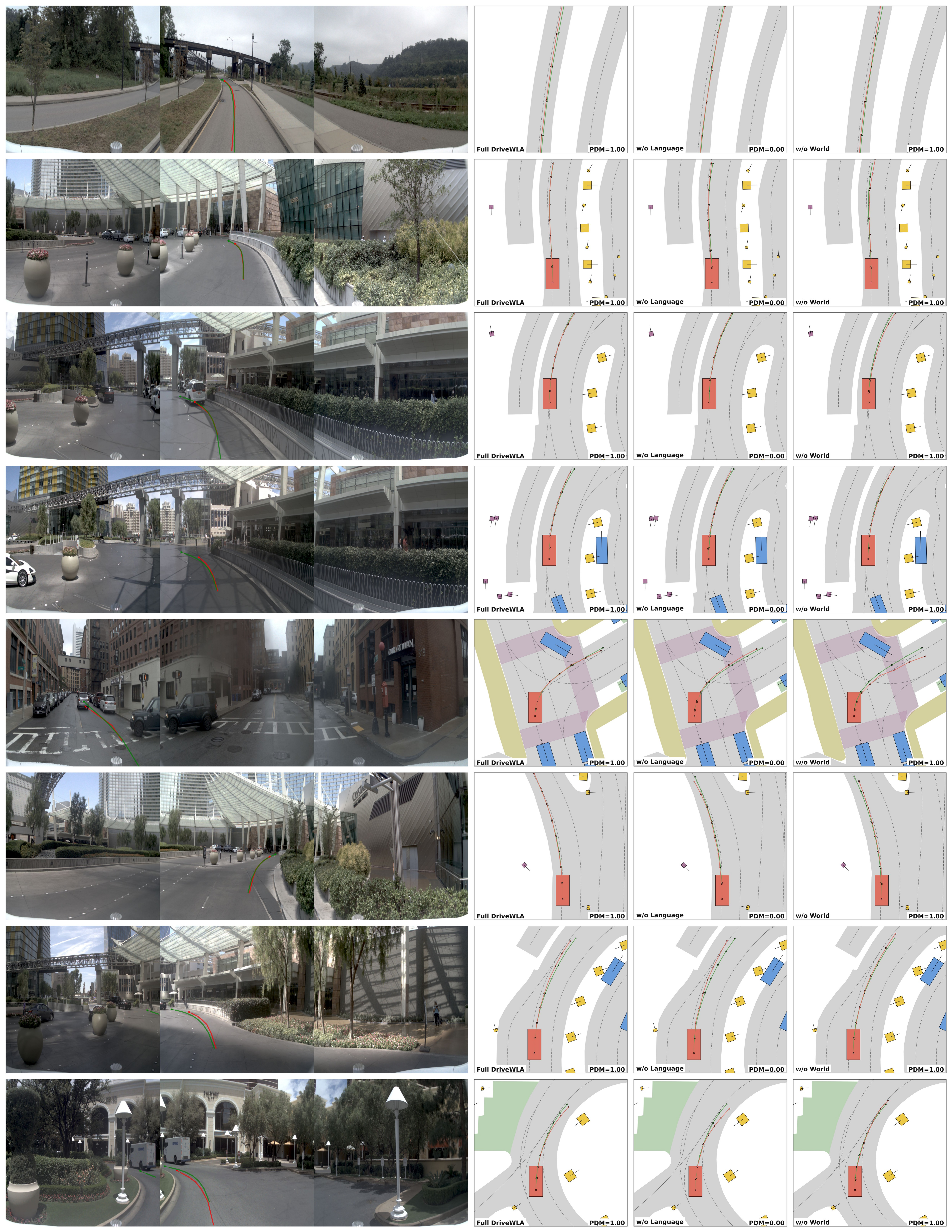}
\caption{\textbf{Language Expert ablation: eight examples from NAVSIM \texttt{navtest}.}
Each row shows a panorama from the left, front, and right cameras for reference, followed by BEV views of Full DriveWLA, w/o Language, and w/o World. Red points are predicted waypoints, and green lines are logged expert trajectories. The three BEV views in each row share the same $26\,\mathrm{m}\times26\,\mathrm{m}$ window. PDM scores range from 0 to 1.}
\label{fig:supp-language-cases}
\end{figure*}

\begin{figure*}[p]
\centering
\includegraphics[width=\textwidth,height=0.89\textheight,keepaspectratio]{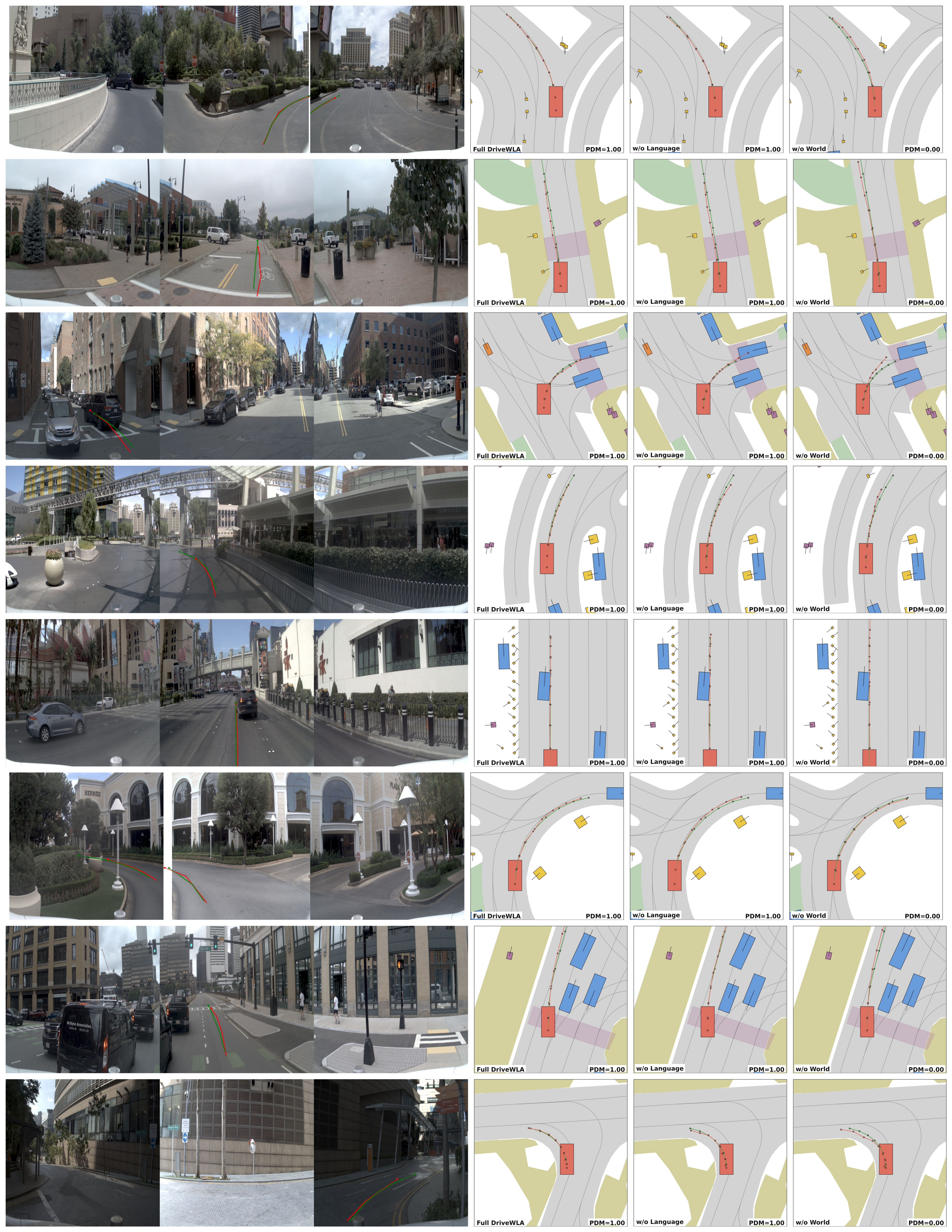}
\caption{\textbf{World Expert ablation: eight examples from NAVSIM \texttt{navtest}.}
Full DriveWLA and w/o Language score 1.00 in each scene, while w/o World scores 0.00. The columns, trajectory colors, BEV windows, and score scale are the same as in Fig.~\ref{fig:supp-language-cases}.}
\label{fig:supp-world-cases}
\end{figure*}

\noindent\textbf{Additional ablation examples.}
Figs.~\ref{fig:supp-language-cases} and~\ref{fig:supp-world-cases} show two sets of eight scenes that highlight the experts' complementary roles. Removing the Language Expert causes failures in the first set, while removing the World Expert causes failures in the second. Both successful models score 1.00 PDM in every scene; the failed model scores 0.00.

\clearpage

\clearpage

\ifdefined\combinedbuild\else
{\small
  \bibliographystyle{ieeenat_fullname}
  \bibliography{main}

\begin{thebibliography}{56}
\providecommand{\natexlab}[1]{#1}
\providecommand{\url}[1]{\texttt{#1}}
\expandafter\ifx\csname urlstyle\endcsname\relax
  \providecommand{\doi}[1]{doi: #1}\else
  \providecommand{\doi}{doi: \begingroup \urlstyle{rm}\Url}\fi

\bibitem[Caesar et~al.(2020)Caesar, Bankiti, Lang, Vora, Liong, Xu, Krishnan, Pan, Baldan, and Beijbom]{caesar2020nuscenes}
Holger Caesar, Varun Bankiti, Alex~H Lang, Sourabh Vora, Venice~Erin Liong, Qiang Xu, Anush Krishnan, Yu Pan, Giancarlo Baldan, and Oscar Beijbom.
\newblock {nuScenes}: A multimodal dataset for autonomous driving.
\newblock In \emph{2020 IEEE/CVF Conference on Computer Vision and Pattern Recognition (CVPR)}, pages 11618--11628. IEEE, 2020.

\bibitem[Cao et~al.(2025)Cao, Hallgarten, Li, Dauner, Gu, Wang, Miron, Aiello, Li, Gilitschenski, et~al.]{cao2025pseudo}
Wei Cao, Marcel Hallgarten, Tianyu Li, Daniel Dauner, Xunjiang Gu, Caojun Wang, Yakov Miron, Marco Aiello, Hongyang Li, Igor Gilitschenski, et~al.
\newblock Pseudo-simulation for autonomous driving.
\newblock \emph{arXiv preprint arXiv:2506.04218}, 2025.

\bibitem[Chitta et~al.(2022)Chitta, Prakash, Jaeger, Yu, Renz, and Geiger]{chitta2022transfuser}
Kashyap Chitta, Aditya Prakash, Bernhard Jaeger, Zehao Yu, Katrin Renz, and Andreas Geiger.
\newblock {TransFuser}: Imitation with transformer-based sensor fusion for autonomous driving.
\newblock \emph{IEEE Transactions on Pattern Analysis and Machine Intelligence}, 45\penalty0 (11):\penalty0 12878--12895, 2022.

\bibitem[Codevilla et~al.(2018)Codevilla, M{\"u}ller, L{\'o}pez, Koltun, and Dosovitskiy]{codevilla2018end}
Felipe Codevilla, Matthias M{\"u}ller, Antonio L{\'o}pez, Vladlen Koltun, and Alexey Dosovitskiy.
\newblock End-to-end driving via conditional imitation learning.
\newblock In \emph{2018 IEEE International Conference on Robotics and Automation (ICRA)}, pages 4693--4700. IEEE, 2018.

\bibitem[Dauner et~al.(2024)Dauner, Hallgarten, Li, Weng, Huang, Yang, Li, Gilitschenski, Ivanovic, Pavone, et~al.]{dauner2024navsim}
Daniel Dauner, Marcel Hallgarten, Tianyu Li, Xinshuo Weng, Zhiyu Huang, Zetong Yang, Hongyang Li, Igor Gilitschenski, Boris Ivanovic, Marco Pavone, et~al.
\newblock {NAVSIM}: Data-driven non-reactive autonomous vehicle simulation and benchmarking.
\newblock \emph{Advances in Neural Information Processing Systems}, 37:\penalty0 28706--28719, 2024.

\bibitem[Engstr{\"o}m et~al.(2018)Engstr{\"o}m, B{\"a}rgman, Nilsson, Seppelt, Markkula, Piccinini, and Victor]{engstrom2018great}
Johan Engstr{\"o}m, Jonas B{\"a}rgman, Daniel Nilsson, Bobbie Seppelt, Gustav Markkula, Giulio~Bianchi Piccinini, and Trent Victor.
\newblock Great expectations: a predictive processing account of automobile driving.
\newblock \emph{Theoretical Issues in Ergonomics Science}, 19\penalty0 (2):\penalty0 156--194, 2018.

\bibitem[Gao et~al.(2026)Gao, Chen, Jiang, Liao, Shi, Guo, Pu, Li, Liu, Zhang, et~al.]{gao2026rad}
Hao Gao, Shaoyu Chen, Bo Jiang, Bencheng Liao, Yiang Shi, Xiaoyang Guo, Yuechuan Pu, Xiangyu Li, Wenyu Liu, Qian Zhang, et~al.
\newblock {RAD}: Training an end-to-end driving policy via large-scale {3DGS}-based reinforcement learning.
\newblock \emph{Advances in Neural Information Processing Systems}, 38:\penalty0 32551--32576, 2026.

\bibitem[HaCohen et~al.(2024)HaCohen, Chiprut, Brazowski, Shalem, Moshe, Richardson, Levin, Shiran, Zabari, Gordon, et~al.]{hacohen2024ltx}
Yoav HaCohen, Nisan Chiprut, Benny Brazowski, Daniel Shalem, Dudu Moshe, Eitan Richardson, Eran Levin, Guy Shiran, Nir Zabari, Ori Gordon, et~al.
\newblock {LTX-Video}: Realtime video latent diffusion.
\newblock \emph{arXiv preprint arXiv:2501.00103}, 2024.

\bibitem[Hong et~al.(2026)Hong, Zhou, Li, Zhou, Liu, Luo, Xu, Yang, and Song]{hong2026drivefuture}
Yufeng Hong, Xiaotian Zhou, Yingyan Li, Xiangpo Zhou, Lin Liu, Yadan Luo, Shaoqing Xu, Lei Yang, and Ziying Song.
\newblock {DriveFuture}: Future-aware latent world models for autonomous driving.
\newblock \emph{arXiv preprint arXiv:2605.09701}, 2026.

\bibitem[Hu et~al.(2023{\natexlab{a}})Hu, Russell, Yeo, Murez, Fedoseev, Kendall, Shotton, and Corrado]{hu2023gaia}
Anthony Hu, Lloyd Russell, Hudson Yeo, Zak Murez, George Fedoseev, Alex Kendall, Jamie Shotton, and Gianluca Corrado.
\newblock {GAIA-1}: A generative world model for autonomous driving.
\newblock \emph{arXiv preprint arXiv:2309.17080}, 2023{\natexlab{a}}.

\bibitem[Hu et~al.(2021)Hu, Shen, Wallis, Allen-Zhu, Li, Wang, Wang, and Chen]{hu2021lora}
Edward~J Hu, Yelong Shen, Phillip Wallis, Zeyuan Allen-Zhu, Yuanzhi Li, Shean Wang, Lu Wang, and Weizhu Chen.
\newblock {LoRA}: Low-rank adaptation of large language models.
\newblock \emph{arXiv preprint arXiv:2106.09685}, 2021.

\bibitem[Hu et~al.(2023{\natexlab{b}})Hu, Yang, Chen, Li, Sima, Zhu, Chai, Du, Lin, Wang, et~al.]{hu2023planning}
Yihan Hu, Jiazhi Yang, Li Chen, Keyu Li, Chonghao Sima, Xizhou Zhu, Siqi Chai, Senyao Du, Tianwei Lin, Wenhai Wang, et~al.
\newblock Planning-oriented autonomous driving.
\newblock In \emph{2023 IEEE/CVF Conference on Computer Vision and Pattern Recognition (CVPR)}, pages 17853--17862. IEEE, 2023{\natexlab{b}}.

\bibitem[Huang et~al.(2026)Huang, Xiang, Liang, Huang, Wang, Xu, Tan, Zhou, Yang, and Che]{huang2026coworld}
Minqing Huang, Yujiao Xiang, Zihan Liang, Jiajie Huang, Jingqi Wang, Zhi Xu, Feiyang Tan, Hangning Zhou, Mu Yang, and Gong Che.
\newblock {CoWorld-VLA}: Thinking in a multi-expert world model for autonomous driving.
\newblock \emph{arXiv preprint arXiv:2605.10426}, 2026.

\bibitem[Huang et~al.(2025)Huang, Weng, Igl, Chen, Cao, Ivanovic, Pavone, and Lv]{huang2025gen}
Zhiyu Huang, Xinshuo Weng, Maximilian Igl, Yuxiao Chen, Yulong Cao, Boris Ivanovic, Marco Pavone, and Chen Lv.
\newblock {Gen-Drive}: Enhancing diffusion generative driving policies with reward modeling and reinforcement learning fine-tuning.
\newblock In \emph{2025 IEEE International Conference on Robotics and Automation (ICRA)}, pages 3445--3451. IEEE, 2025.

\bibitem[Hwang et~al.(2024)Hwang, Xu, Lin, Hung, Ji, Choi, Huang, He, Covington, Sapp, et~al.]{hwang2024emma}
Jyh-Jing Hwang, Runsheng Xu, Hubert Lin, Wei-Chih Hung, Jingwei Ji, Kristy Choi, Di Huang, Tong He, Paul Covington, Benjamin Sapp, et~al.
\newblock {EMMA}: End-to-end multimodal model for autonomous driving.
\newblock \emph{arXiv preprint arXiv:2410.23262}, 2024.

\bibitem[Jia et~al.(2024)Jia, Yang, Li, Zhang, and Yan]{jia2024bench2drive}
Xiaosong Jia, Zhenjie Yang, Qifeng Li, Zhiyuan Zhang, and Junchi Yan.
\newblock {Bench2Drive}: Towards multi-ability benchmarking of closed-loop end-to-end autonomous driving.
\newblock \emph{Advances in Neural Information Processing Systems}, 37:\penalty0 819--844, 2024.

\bibitem[Jiang et~al.(2023)Jiang, Chen, Xu, Liao, Chen, Zhou, Zhang, Liu, Huang, and Wang]{jiang2023vad}
Bo Jiang, Shaoyu Chen, Qing Xu, Bencheng Liao, Jiajie Chen, Helong Zhou, Qian Zhang, Wenyu Liu, Chang Huang, and Xinggang Wang.
\newblock {VAD}: Vectorized scene representation for efficient autonomous driving.
\newblock In \emph{2023 IEEE/CVF International Conference on Computer Vision (ICCV)}, pages 8306--8316. IEEE, 2023.

\bibitem[Kirby et~al.(2026)Kirby, Boulch, Xu, Yin, Puy, Zablocki, Bursuc, Gidaris, Marlet, Bartoccioni, et~al.]{kirby2026driving}
Ellington Kirby, Alexandre Boulch, Yihong Xu, Yuan Yin, Gilles Puy, {\'E}loi Zablocki, Andrei Bursuc, Spyros Gidaris, Renaud Marlet, Florent Bartoccioni, et~al.
\newblock Driving on registers.
\newblock In \emph{Proceedings of the IEEE/CVF Conference on Computer Vision and Pattern Recognition}, pages 32058--32069, 2026.

\bibitem[Li et~al.(2026{\natexlab{a}})Li, Wu, Hu, Huang, Sun, Hao, Lang, Zhu, and Zhang]{li2026sgdrive}
Jingyu Li, Junjie Wu, Dongnan Hu, Xiangkai Huang, Bin Sun, Zhihui Hao, Xianpeng Lang, Xiatian Zhu, and Li Zhang.
\newblock {SGDrive}: Scene-to-goal hierarchical world cognition for autonomous driving.
\newblock \emph{arXiv preprint arXiv:2601.05640}, 2026{\natexlab{a}}.

\bibitem[Li et~al.(2026{\natexlab{b}})Li, Shang, Liu, Zhan, Wang, Wang, Chen, Wang, An, Tang, et~al.]{li2026drivevla}
Yingyan Li, Shuyao Shang, Weisong Liu, Bing Zhan, Haochen Wang, Yuqi Wang, Yuntao Chen, Xiaoman Wang, Yasong An, Chufeng Tang, et~al.
\newblock {DriveVLA-W0}: World models amplify data scaling law in autonomous driving.
\newblock In \emph{International Conference on Learning Representations}, pages 7890--7911, 2026{\natexlab{b}}.

\bibitem[Li et~al.(2026{\natexlab{c}})Li, Zhou, Yan, Liao, Yan, Xiong, Chen, Xie, Wang, Chen, et~al.]{li2026unidrivevla}
Yongkang Li, Lijun Zhou, Sixu Yan, Bencheng Liao, Tianyi Yan, Kaixin Xiong, Long Chen, Hongwei Xie, Bing Wang, Guang Chen, et~al.
\newblock {UniDriveVLA}: Unifying understanding, perception, and action planning for autonomous driving.
\newblock \emph{arXiv preprint arXiv:2604.02190}, 2026{\natexlab{c}}.

\bibitem[Liang et~al.(2024)Liang, Yu, Luo, Iyer, Dong, Zhou, Ghosh, Lewis, Yih, Zettlemoyer, et~al.]{liang2024mixture}
Weixin Liang, Lili Yu, Liang Luo, Srinivasan Iyer, Ning Dong, Chunting Zhou, Gargi Ghosh, Mike Lewis, Wen-tau Yih, Luke Zettlemoyer, et~al.
\newblock Mixture-of-transformers: A sparse and scalable architecture for multi-modal foundation models.
\newblock \emph{arXiv preprint arXiv:2411.04996}, 2024.

\bibitem[Liao et~al.(2025)Liao, Chen, Yin, Jiang, Wang, Yan, Zhang, Li, Zhang, Zhang, et~al.]{liao2025diffusiondrive}
Bencheng Liao, Shaoyu Chen, Haoran Yin, Bo Jiang, Cheng Wang, Sixu Yan, Xinbang Zhang, Xiangyu Li, Ying Zhang, Qian Zhang, et~al.
\newblock {DiffusionDrive}: Truncated diffusion model for end-to-end autonomous driving.
\newblock In \emph{2025 IEEE/CVF Conference on Computer Vision and Pattern Recognition (CVPR)}, pages 12037--12047. IEEE, 2025.

\bibitem[Liao et~al.(2022)Liao, Xie, and Geiger]{liao2022kitti}
Yiyi Liao, Jun Xie, and Andreas Geiger.
\newblock {KITTI-360}: A novel dataset and benchmarks for urban scene understanding in {2D} and {3D}.
\newblock \emph{IEEE Transactions on Pattern Analysis and Machine Intelligence}, 45\penalty0 (3):\penalty0 3292--3310, 2022.

\bibitem[Lipman et~al.(2022)Lipman, Chen, Ben-Hamu, Nickel, and Le]{lipman2022flow}
Yaron Lipman, Ricky~TQ Chen, Heli Ben-Hamu, Maximilian Nickel, and Matt Le.
\newblock Flow matching for generative modeling.
\newblock \emph{arXiv preprint arXiv:2210.02747}, 2022.

\bibitem[Liu et~al.(2026{\natexlab{a}})Liu, Liu, Liang, Li, Liu, Wang, Wan, Zhang, and Ouyang]{liu2026flow}
Jie Liu, Gongye Liu, Jiajun Liang, Yangguang Li, Jiaheng Liu, Xintao Wang, Pengfei Wan, Di Zhang, and Wanli Ouyang.
\newblock {Flow-GRPO}: Training flow matching models via online {RL}.
\newblock \emph{Advances in Neural Information Processing Systems}, 38:\penalty0 40783--40818, 2026{\natexlab{a}}.

\bibitem[Liu et~al.(2026{\natexlab{b}})Liu, Song, Jia, Ye, Hao, Chen, et~al.]{liu2026driveworld}
Lin Liu, Ziying Song, Caiyan Jia, Hangjun Ye, Xiaoshuai Hao, Long Chen, et~al.
\newblock {DriveWorld-VLA}: Unified latent-space world modeling with vision-language-action for autonomous driving.
\newblock \emph{arXiv preprint arXiv:2602.06521}, 2026{\natexlab{b}}.

\bibitem[Liu et~al.(2026{\natexlab{c}})Liu, Zhang, Liu, Cui, Xie, Chen, Ye, Nex, Cheng, and Yang]{liu2026universe}
Mengmeng Liu, Diankun Zhang, Jiuming Liu, Jianfeng Cui, Hongwei Xie, Guang Chen, Hangjun Ye, Francesco Nex, Hao Cheng, and Michael~Ying Yang.
\newblock {UNIVERSE}: Unified video action models for autonomous driving with flexible mask-modulated modality generation.
\newblock \emph{arXiv preprint arXiv:2607.05133}, 2026{\natexlab{c}}.

\bibitem[Liu et~al.(2026{\natexlab{d}})Liu, Zhang, Liu, Cui, Xie, Chen, Ye, Yang, Nex, and Cheng]{liu2026driveva}
Mengmeng Liu, Diankun Zhang, Jiuming Liu, Jianfeng Cui, Hongwei Xie, Guang Chen, Hangjun Ye, Michael~Ying Yang, Francesco Nex, and Hao Cheng.
\newblock {DriveVA}: Video action models are zero-shot drivers.
\newblock In \emph{European Conference on Computer Vision}, pages 315--335. Springer, 2026{\natexlab{d}}.

\bibitem[Peebles and Xie(2023)]{peebles2023scalable}
William Peebles and Saining Xie.
\newblock Scalable diffusion models with transformers.
\newblock In \emph{2023 IEEE/CVF International Conference on Computer Vision (ICCV)}, pages 4172--4182. IEEE, 2023.

\bibitem[Shao et~al.(2024)Shao, Hu, Wang, Song, Waslander, Liu, and Li]{shao2024lmdrive}
Hao Shao, Yuxuan Hu, Letian Wang, Guanglu Song, Steven~L Waslander, Yu Liu, and Hongsheng Li.
\newblock {LMDrive}: Closed-loop end-to-end driving with large language models.
\newblock In \emph{2024 IEEE/CVF Conference on Computer Vision and Pattern Recognition (CVPR)}, pages 15120--15130. IEEE, 2024.

\bibitem[Shi et~al.(2026)Shi, Xu, Shi, Sheng, Zhang, and Jiang]{shi2026drivewam}
Chen Shi, Jinrui Xu, Shaoshuai Shi, Kehua Sheng, Bo Zhang, and Li Jiang.
\newblock {DriveWAM}: Video generative priors enable scalable world-action modeling for autonomous driving.
\newblock \emph{arXiv preprint arXiv:2605.28544}, 2026.

\bibitem[Sun et~al.(2020)Sun, Kretzschmar, Dotiwalla, Chouard, Patnaik, Tsui, Guo, Zhou, Chai, Caine, et~al.]{sun2020scalability}
Pei Sun, Henrik Kretzschmar, Xerxes Dotiwalla, Aurelien Chouard, Vijaysai Patnaik, Paul Tsui, James Guo, Yin Zhou, Yuning Chai, Benjamin Caine, et~al.
\newblock Scalability in perception for autonomous driving: {Waymo} open dataset.
\newblock In \emph{2020 IEEE/CVF Conference on Computer Vision and Pattern Recognition (CVPR)}, pages 2443--2451. IEEE, 2020.

\bibitem[Tian et~al.(2024)Tian, Gu, Li, Liu, Wang, Zhao, Zhan, Jia, Lang, and Zhao]{tian2024drivevlm}
Xiaoyu Tian, Junru Gu, Bailin Li, Yicheng Liu, Yang Wang, Zhiyong Zhao, Kun Zhan, Peng Jia, Xianpeng Lang, and Hang Zhao.
\newblock {DriveVLM}: The convergence of autonomous driving and large vision-language models.
\newblock \emph{arXiv preprint arXiv:2402.12289}, 2024.

\bibitem[Wang et~al.(2026{\natexlab{a}})Wang, Zheng, Chen, Li, Zhang, Xing, Zhang, Li, Qian, Yang, et~al.]{wang2026latent}
Linbo Wang, Yupeng Zheng, Qiang Chen, Shiwei Li, Yichen Zhang, Zebin Xing, Qichao Zhang, Xiang Li, Deheng Qian, Pengxuan Yang, et~al.
\newblock {Latent-WAM}: Latent world action modeling for end-to-end autonomous driving.
\newblock \emph{arXiv preprint arXiv:2603.24581}, 2026{\natexlab{a}}.

\bibitem[Wang et~al.(2024{\natexlab{a}})Wang, Zhu, Huang, Chen, Zhu, and Lu]{wang2024drivedreamer}
Xiaofeng Wang, Zheng Zhu, Guan Huang, Xinze Chen, Jiagang Zhu, and Jiwen Lu.
\newblock {DriveDreamer}: Towards real-world-drive world models for autonomous driving.
\newblock In \emph{European Conference on Computer Vision}, pages 55--72. Springer, 2024{\natexlab{a}}.

\bibitem[Wang et~al.(2026{\natexlab{b}})Wang, Xiang, Zhou, Wang, Huang, Huang, Wei, Zhou, Wang, Chen, et~al.]{wang2026wa}
Xinlin Wang, Yujiao Xiang, Yuheng Zhou, Jingqi Wang, Minqing Huang, Jiajie Huang, Dongxu Wei, Tingguang Zhou, Xiyang Wang, Gong Chen, et~al.
\newblock {WA-JEPA}: Rethinking the video {JEPA} paradigm for world-action modeling in autonomous driving.
\newblock \emph{arXiv preprint arXiv:2608.20974}, 2026{\natexlab{b}}.

\bibitem[Wang et~al.(2024{\natexlab{b}})Wang, He, Fan, Li, Chen, and Zhang]{wang2024driving}
Yuqi Wang, Jiawei He, Lue Fan, Hongxin Li, Yuntao Chen, and Zhaoxiang Zhang.
\newblock Driving into the future: Multiview visual forecasting and planning with world model for autonomous driving.
\newblock In \emph{2024 IEEE/CVF Conference on Computer Vision and Pattern Recognition (CVPR)}, pages 14749--14759. IEEE, 2024{\natexlab{b}}.

\bibitem[Wolfe et~al.(2019)Wolfe, Fridman, Kosovicheva, Seppelt, Mehler, Reimer, and Rosenholtz]{wolfe2019predicting}
Benjamin Wolfe, Lex Fridman, Anna Kosovicheva, Bobbie Seppelt, Bruce Mehler, Bryan Reimer, and Ruth Rosenholtz.
\newblock Predicting road scenes from brief views of driving video.
\newblock \emph{Journal of Vision}, 19\penalty0 (5):\penalty0 8--8, 2019.

\bibitem[Xia et~al.(2026{\natexlab{a}})Xia, Li, Zhou, Yao, Xiong, Sun, Wang, Ma, Chen, Ye, et~al.]{xia2026drivelaw}
Tianze Xia, Yongkang Li, Lijun Zhou, Jingfeng Yao, Kaixin Xiong, Haiyang Sun, Bing Wang, Kun Ma, Guang Chen, Hangjun Ye, et~al.
\newblock {DriveLaW}: Unifying planning and video generation in a latent driving world.
\newblock In \emph{Proceedings of the IEEE/CVF Conference on Computer Vision and Pattern Recognition}, pages 39701--39712, 2026{\natexlab{a}}.

\bibitem[Xia et~al.(2026{\natexlab{b}})Xia, Zhou, Xiong, Yao, Zhu, Zhu, Wang, Chen, Ye, Liu, et~al.]{xia2026reworld}
Tianze Xia, Lijun Zhou, Kaixin Xiong, Jingfeng Yao, Yu Zhu, Zhenxin Zhu, Bing Wang, Guang Chen, Hangjun Ye, Wenyu Liu, et~al.
\newblock {ReWorld}: Learning better representations for world action models.
\newblock \emph{arXiv preprint arXiv:2606.27504}, 2026{\natexlab{b}}.

\bibitem[Xiao et~al.(2021)Xiao, Shao, Hao, Zhang, Chai, Jiao, Li, Wu, Sun, Jiang, et~al.]{xiao2021pandaset}
Pengchuan Xiao, Zhenlei Shao, Steven Hao, Zishuo Zhang, Xiaolin Chai, Judy Jiao, Zesong Li, Jian Wu, Kai Sun, Kun Jiang, et~al.
\newblock {PandaSet}: Advanced sensor suite dataset for autonomous driving.
\newblock In \emph{2021 IEEE International Intelligent Transportation Systems Conference (ITSC)}, pages 3095--3101. IEEE, 2021.

\bibitem[Xiong et~al.(2026)Xiong, Guo, Li, Yan, Xu, Zhou, Chen, Sun, Wang, Ma, et~al.]{xiong2026recogdrive}
Kaixin Xiong, Xiangyu Guo, Fang Li, Sixu Yan, Gangwei Xu, Lijun Zhou, Long Chen, Haiyang Sun, Bing Wang, Kun Ma, et~al.
\newblock {ReCogDrive}: A reinforced cognitive framework for end-to-end autonomous driving.
\newblock In \emph{International Conference on Learning Representations}, pages 157518--157556, 2026.

\bibitem[Xu et~al.(2026)Xu, Zhong, Shu, Jia, Li, Bian, Zhang, Zhang, Xie, Yang, et~al.]{xu2026eponav2}
Jiawei Xu, Zhizhou Zhong, Zhijian Shu, Mingkai Jia, Mingxiao Li, Jia-Wang Bian, Qian Zhang, Kaicheng Zhang, Jin Xie, Jian Yang, et~al.
\newblock {EponaV2}: Driving world model with comprehensive future reasoning.
\newblock \emph{arXiv preprint arXiv:2605.14696}, 2026.

\bibitem[Yang et~al.(2026)Yang, Jia, Li, Yang, Yao, and Yan]{yang2026raw2drive}
Zhenjie Yang, Xiaosong Jia, Qifeng Li, Xue Yang, Maoqing Yao, and Junchi Yan.
\newblock {Raw2Drive}: Reinforcement learning with aligned world models for end-to-end autonomous driving (in {CARLA} v2).
\newblock \emph{Advances in Neural Information Processing Systems}, 38:\penalty0 134122--134147, 2026.

\bibitem[Zhan et~al.(2026)Zhan, Shang, Gu, Lu, Xu, Wang, Wang, Zhang, Zhan, Fan, et~al.]{zhan2026brainwam}
Bing Zhan, Shuyao Shang, Jiahao Gu, Shuo Lu, Yuan Xu, Zhao Wang, Yida Wang, Xueyang Zhang, Kun Zhan, Lue Fan, et~al.
\newblock {BrainWAM}: Action-space coordination of semantic priors and predictive dynamics for autonomous driving.
\newblock \emph{arXiv preprint arXiv:2608.12854}, 2026.

\bibitem[Zhang et~al.(2025)Zhang, Liang, Guo, Lu, Wang, Xiong, Miao, and Wang]{zhang2025carplanner}
Dongkun Zhang, Jiaming Liang, Ke Guo, Sha Lu, Qi Wang, Rong Xiong, Zhenwei Miao, and Yue Wang.
\newblock {CarPlanner}: Consistent auto-regressive trajectory planning for large-scale reinforcement learning in autonomous driving.
\newblock In \emph{2025 IEEE/CVF Conference on Computer Vision and Pattern Recognition (CVPR)}, pages 17239--17248. IEEE, 2025.

\bibitem[Zhao et~al.(2025)Zhao, Wang, Zhu, Chen, Huang, Bao, and Wang]{zhao2025drivedreamer}
Guosheng Zhao, Xiaofeng Wang, Zheng Zhu, Xinze Chen, Guan Huang, Xiaoyi Bao, and Xingang Wang.
\newblock {DriveDreamer-2}: {LLM}-enhanced world models for diverse driving video generation.
\newblock In \emph{Proceedings of the AAAI Conference on Artificial Intelligence}, pages 10412--10420, 2025.

\bibitem[Zhao et~al.(2026{\natexlab{a}})Zhao, Fu, Wang, Wang, and Lu]{zhao2026forecasting}
Zhida Zhao, Talas Fu, Yifan Wang, Lijun Wang, and Huchuan Lu.
\newblock From forecasting to planning: Policy world model for collaborative state-action prediction.
\newblock \emph{Advances in Neural Information Processing Systems}, 38:\penalty0 134585--134611, 2026{\natexlab{a}}.

\bibitem[Zhao et~al.(2026{\natexlab{b}})Zhao, Zhou, Xu, Sun, Zhou, Li, Liang, and Bai]{zhao2026simwam}
Zongchuang Zhao, Xin Zhou, Tianyang Xu, Zhengyang Sun, Kaixuan Zhou, Honglin Li, Dingkang Liang, and Xiang Bai.
\newblock {SimWAM}: A simple world action model for end-to-end autonomous driving.
\newblock \emph{arXiv preprint arXiv:2608.07468}, 2026{\natexlab{b}}.

\bibitem[Zhou et~al.(2025)Zhou, Lin, Wang, Lu, Bai, Liu, Wang, Geiger, and Liao]{zhou2025hugsim}
Hongyu Zhou, Longzhong Lin, Jiabao Wang, Yichong Lu, Dongfeng Bai, Bingbing Liu, Yue Wang, Andreas Geiger, and Yiyi Liao.
\newblock {HUGSIM}: A real-time, photo-realistic and closed-loop simulator for autonomous driving.
\newblock \emph{IEEE Transactions on Pattern Analysis and Machine Intelligence}, 2025.

\bibitem[Zhou et~al.(2026{\natexlab{a}})Zhou, Han, Yang, Ma, Tresp, and Knoll]{zhou2026opendrivevla}
Xingcheng Zhou, Xuyuan Han, Feng Yang, Yunpu Ma, Volker Tresp, and Alois Knoll.
\newblock {OpenDriveVLA}: Towards end-to-end autonomous driving with large vision language action model.
\newblock In \emph{Proceedings of the AAAI Conference on Artificial Intelligence}, pages 13782--13790, 2026{\natexlab{a}}.

\bibitem[Zhou et~al.(2026{\natexlab{b}})Zhou, Zhao, Yang, Li, Zhong, Bai, Chu, Chen, Li, Tang, et~al.]{zhou2026qwen}
Xin Zhou, Zongchuang Zhao, Zhibo Yang, Mingsheng Li, Humen Zhong, Shuai Bai, Du Chu, Ruizhe Chen, Zhaohai Li, Jun Tang, et~al.
\newblock {Qwen-Drive-1.0}: An initial step towards a vision-language foundation model for autonomous driving.
\newblock \emph{arXiv preprint arXiv:2609.00111}, 2026{\natexlab{b}}.

\bibitem[Zhou et~al.(2026{\natexlab{c}})Zhou, Cai, Zhao, Zhang, Huang, Zhou, and Ma]{zhou2026autovla}
Zewei Zhou, Tianhui Cai, Seth Zhao, Yun Zhang, Zhiyu Huang, Bolei Zhou, and Jiaqi Ma.
\newblock {AutoVLA}: A vision-language-action model for end-to-end autonomous driving with adaptive reasoning and reinforcement fine-tuning.
\newblock \emph{Advances in Neural Information Processing Systems}, 38:\penalty0 27920--27956, 2026{\natexlab{c}}.

\bibitem[Zhu et~al.(2025)Zhu, Wang, Chen, Liu, Ye, Gu, Tian, Duan, Su, Shao, et~al.]{zhu2025internvl3}
Jinguo Zhu, Weiyun Wang, Zhe Chen, Zhaoyang Liu, Shenglong Ye, Lixin Gu, Hao Tian, Yuchen Duan, Weijie Su, Jie Shao, et~al.
\newblock {InternVL3}: Exploring advanced training and test-time recipes for open-source multimodal models.
\newblock \emph{arXiv preprint arXiv:2504.10479}, 2025.

\bibitem[Zou et~al.(2025)Zou, Chen, Liao, Zheng, Song, Zhang, Zhang, Liu, and Wang]{zou2025diffusiondrivev2}
Jialv Zou, Shaoyu Chen, Bencheng Liao, Zhiyu Zheng, Yuehao Song, Lefei Zhang, Qian Zhang, Wenyu Liu, and Xinggang Wang.
\newblock {DiffusionDriveV2}: Reinforcement learning-constrained truncated diffusion modeling in end-to-end autonomous driving.
\newblock \emph{arXiv preprint arXiv:2512.07745}, 2025.

\end{thebibliography}
}
\fi

\end{document}